\documentclass{article}

\usepackage[preprint]{neurips_2025}

\usepackage{amsmath}
\usepackage{amsfonts}

\usepackage{pifont}

\def\va{{\mathbf{a}}}

\def\sR{\mathbb{R}}

\def\sN{\mathbb{N}}

\def\va{{\mathbf{a}}}
\def\vb{{\mathbf{b}}}
\def\vc{{\mathbf{c}}}

\def\vx{{\mathbf{x}}}
\def\vy{{\mathbf{y}}}

\def\mA{{\mathbf{A}}}
\def\mB{{\mathbf{B}}}

\def\mW{{\mathbf{W}}}
\def\mX{{\mathbf{X}}}

\def\eE{\mathbb{E}}

\def\pP{\mathbb{P}}

\def\defeq{\overset{\textup{def}}{=}}

\def\leqone{\overset{\text{\ding{172}}}{\leq}}
\def\leqtwo{\overset{\text{\ding{173}}}{\leq}}
\def\leqthree{\overset{\text{\ding{174}}}{\leq}}
\def\leqfour{\overset{\text{\ding{175}}}{\leq}}
\def\eqone{\overset{\text{\ding{172}}}{=}}
\def\eqtwo{\overset{\text{\ding{173}}}{=}}
\def\eqthree{\overset{\text{\ding{174}}}{=}}
\def\eqfour{\overset{\text{\ding{175}}}{=}}
\def\geqfive{\overset{\text{\ding{176}}}{\geq}}

\usepackage[utf8]{inputenc} 
\usepackage[T1]{fontenc}    
\usepackage{hyperref}       
\usepackage{url}            
\usepackage{booktabs}       
\usepackage{amsfonts}       
\usepackage{nicefrac}       
\usepackage{microtype}      
\usepackage{xcolor}         

\usepackage{tikz}

\usepackage{color-edits}
\addauthor{jm}{purple}
\addauthor{pa}{green}
\addauthor{mm}{orange}
\addauthor{cc}{red}
\usepackage{mathtools}
\usepackage{amsthm}
\usepackage{graphicx}
\usepackage{enumerate}
\usepackage{algorithm}
\usepackage{algorithmic}
\usepackage{thm-restate}

\usepackage{url}
\usepackage{dsfont}
\usepackage[mathscr]{euscript}
\usepackage{booktabs}
\usepackage{makecell}
\usepackage{tabularx}
\usepackage{xcolor, colortbl}
\usepackage[normalem]{ulem}
\usepackage{soul}
\usepackage{prettyref}

\usepackage{MnSymbol}
\DeclareMathAlphabet\mathbb{U}{msb}{m}{n}
\usepackage{xpatch}

\DeclareMathOperator*{\argmin}{argmin}

\newtheorem{theorem}{Theorem}
\newtheorem{assumption}{Assumption}
\newtheorem{proposition}{Proposition}
\newtheorem{corollary}{Corollary}
\newtheorem{lemma}{Lemma}
\newtheorem{definition}{Definition}
\newtheorem*{remark}{Remark}

\hypersetup{
  colorlinks   = true,
  urlcolor     = blue,
  linkcolor    = blue,
  citecolor    = blue
}

\usepackage[textsize=tiny]{todonotes}

\usepackage[toc, page, header]{appendix}
\usepackage{threeparttable}
\usepackage{cleveref}
\title{Sharper Risk Bound for Multi-Task Learning with Multi-Graph Dependent Data}
\author{%
   Xiao~Shao\thanks{Personal homepage: \url{https://time.sdu.edu.cn/info/1069/2646.htm}; Email address: \texttt{xiaoserendipity@mail.sdu.edu.cn}} \\
  School of Software \\ 
  Shandong University \\
  No. 17923, Jingshi Road, Jinan 250061, China  \\
  \texttt{xiaoserendipity@mail.sdu.edu.cn}\\ 
  \And
  Guoqiang~Wu\thanks{Personal homepage: \url{https://guoqiangwoodrowwu.github.io/}; Email address: \texttt{guoqiangwu90@gmail.com}} \\
  School of Software \\
  Shandong University \\
  No. 17923, Jingshi Road, Jinan 250061, China  \\
  \texttt{guoqiangwu90@gmail.com}
}

\begin{document}

\maketitle

\begin{abstract}
  In multi-task learning (MTL) with each task involving graph-dependent data,
existing generalization analyses yield a \emph{sub-optimal} risk bound of $O(\frac{1}{\sqrt{n}})$, where $n$ is the number of training samples of each task. However, to improve the risk bound is technically challenging, which is attributed to the lack of a foundational sharper concentration inequality for multi-graph dependent random variables. 
To fill up this gap, this paper proposes a new Bennett-type inequality, enabling the derivation of a sharper risk bound of $O(\frac{\log n}{n})$. 
Technically, building on the proposed Bennett-type inequality, we propose a new Talagrand-type inequality for the empirical process, and further develop a new analytical framework of the local fractional Rademacher complexity to enhance generalization analyses in MTL with multi-graph dependent data. 
Finally, we apply the theoretical advancements to applications such as Macro-AUC optimization, illustrating the superiority of our theoretical results over prior work, which is also verified by experimental results.
\end{abstract}

\section{Introduction}
\label{sec:introduction}

As a crucial learning task, Multi-Task Learning (MTL)~\cite{caruana1997multitask,zhang2021survey}, has recently attracted significant attention in various fields, including natural language processing~\cite{liu2019multi,chen2024multi} and computer vision~\cite{wong2023fast}.

Theoretically, most studies have made significant progress in generalization analyses within the context of MTL under the assumption that the data of each task is independently and identically distributed (i.i.d.). For example, \cite{baxter95,maurer06linear,maurer2016benefit} analyzed the generalization bounds by utilizing the Rademacher Complexity (RC) \cite{rademacher02initial} and Covering Number (CN) \cite{covering02initial}, resulting in a risk bound of $O(\frac{1}{\sqrt{n}})$, where $n$ is the number of training samples of each task. To achieve tighter bounds, \cite{yousefi18,watkins2023optimistic} employed the Local Rademacher Complexity (LRC), which can produce a better result of $O(\frac{\log n}{n})$.

However, in practical learning scenarios of MTL, there are some non-i.i.d. situations, particularly when the data of each task exhibits a graph-dependent structure~\cite{wu2022mtgcn,21multiclass,wu2023macro-auc},
referred to as \emph{multi-graph dependence}, which is the focus of this paper. (See related work of other dependent structure cases in Appendix~\ref{section: D}). Recently, \cite{wu2023macro-auc} (Appendix~A therein) have conducted generalization analyses of MTL with multi-graph dependent data, resulting in a risk bound of $O(\frac{1}{\sqrt{n}})$
Nevertheless, this result is comparatively weaker than that achieved under the i.i.d. scenario. Naturally, a question then arises:

\quad \textbf{Q1}: \emph{Can we obtain a better risk bound than $O(\frac{1}{\sqrt{n}})$ in MTL with multi-graph dependent data?}

From a technical standpoint, answering the above question is quite challenging, as it necessitates revisiting the foundational tools of risk bounds in learning theory~\cite{99learningtheory,shalev2014understanding,mohri2018foundations}, specifically the concentration inequality~\cite{concentration13initial}, in the context of (multi-)graph dependent random variables.
For single-task learning, \cite{jason04,usunier2005generalization,zhang19mcdiarmid,ralaivola2015entropy} proposed versions of the Hoeffding, McDiarmid, and Bennett concentration inequalities for different function forms of single-graph dependent random variables~\cite{zhang2022generalization}, respectively. 
In contrast, for MTL,~\cite{wu2023macro-auc}~proposed a McDiarmid-type concentration inequality for multi-graph dependent variables. Then, a natural fundamental question arises:

\quad \textbf{Q2}: \emph{Can we develop a new Bennett-type inequality for multi-graph dependent variables?}

This paper provides an affirmative answer to the above question \textbf{Q2} and subsequently addresses the question \textbf{Q1} affirmatively. 
See the following summary of contributions for details.

\textbf{Contributions}. Our main contributions are summarized in Table~\ref{table: technique_results_discuss}, which compares with related work regarding main results and techniques. Specifically, the main contributions are as follows.
\begin{enumerate}[1.]
\setlength\itemsep{-0.2pt}
\vspace{-0.1in}
    \item \textbf{New concentration inequalities.} Technically, we propose a new Bennett-type concentration inequality (Theorem \ref{thm:bennett_inequality}) for multi-graph dependent variables, which can cover the result of single-graph dependent variables~\cite{ralaivola2015entropy} as a special case.\footnote{Notably, a special case of the Bennett-type inequality we propose (Theorem~\ref{thm:bennett_inequality_refined}) can encompass the one in~\cite{Bartlett_2005} under the i.i.d. case, while prior work~\cite{ralaivola2015entropy} cannot.}Further, we propose new corresponding Talagrand-type concentration inequalities for the empirical process (Theorem \ref{thm:talagrand_inequality} and Corollary~\ref{thm:talagrand_inequality_refined}). 
    \item \textbf{Fast rate via a new analytical framework of LFRC.} Based on new concentration inequalities, we develop a new analytical framework of local fractional Rademacher complexity (LFRC), including fast-rate techniques of the sub-foot function and fixed point, for generalization analyses of MTL with multi-graph dependent data, obtaining a fast rate of $O(\frac{\log n}{n})$ over $O(\frac{1}{\sqrt{n}})$ in~\cite{wu2023macro-auc}.
    \item \textbf{Applying to analyze Macro-AUC with experiment verification.} We apply the theoretical advances to analyze the problem of Macro-AUC optimization (see other applications in Appendix~\ref{section: E}). 
    Notably, for this problem, we can obtain a sharper risk bound of $O(\frac{\log \tilde{n}}{\tilde{n}})$ than the previous one of $O(\frac{1}{\sqrt{\tilde{n}}})$ in~\cite{wu2023macro-auc}, where $\tilde{n}$ is the size of training samples, indicating the superiority of our theory results, which is also verified by experimental results.
\end{enumerate}

\begin{table}[t]
\caption{Summary of our contributions and comparison with related work regarding main results and techniques. Our contributions are labelled as red.}
\label{table: technique_results_discuss}
\begin{small}
\centering
\begin{threeparttable} 
\centering
\begin{tabular}
{p{0.5cm}p{4.5cm}p{1.3cm}p{1.5cm}p{1.5cm}p{1.2cm}p{1.5cm}}
\toprule
Ref. & Setting & Convergence Rate & Bennett-type inequality & Talagrand-type inequality & Analog of Thm. 2.1 in \cite{Bartlett_2005} & Fast-rate technique (sub-root function, fixed point) \\
\midrule
 \cite{Bartlett_2005} &  single-task, i.i.d. & $O(\frac{\log n}{n})$ &  -- & --  & \ding{51} & \ding{51} \\
 \cite{ralaivola2015entropy} & single-task,  single-graph dependent  & -- & \ding{51} & \ding{51} & \ding{51} & -- \\
 \cite{yousefi18} & multi-task, i.i.d. & $O(\frac{\log n}{n})$ & -- & -- & \ding{51} & \ding{51} \\
 \cite{watkins2023optimistic} & multi-task representation, i.i.d. & $O(\frac{\log n}{n})$ & -- & \ding{51} & \ding{51} & \ding{51} \\
 \cite{wu2023macro-auc} & multi-task, multi-graph dependent & $O(\frac{1}{\sqrt{n}})$ &  --\tnote{1} & -- & -- &-- \\
 $\color{red} \text{Ours}$ & multi-task, multi-graph dependent & $\color{red} O(\frac{\log n}{n})$ & $\color{red} \text{\ding{51}}$ & $\color{red} \text{\ding{51}}$ & $\color{red} \text{\ding{51}}$ & $\color{red} \text{\ding{51}}$ \\
\bottomrule
\end{tabular}
\tiny
\begin{tablenotes}
    \footnotesize
    \item[1] In fact, it uses the McDiarmid-type inequality and fractional (global) Rademacher complexity.
\end{tablenotes}
\end{threeparttable}
\end{small}
\end{table}

\subsection{Related Work}
\textbf{Concentration inequalities.}
Concentration inequalities \cite{concentration13initial} are important in both statistical learning theory~\cite{statistics97} and probability theory~\cite{07probability}. Here we mainly review the ones of graph-dependent random variables. A methodology analogous to the Hoeffding inequality was introduced to address the summation of graph-dependent variables~\cite{jason04}. Based on this work, \cite{usunier2005generalization} and~\cite{ralaivola2015entropy} proposed new concentration inequalities that both extended the results of \cite{jason04} and broadened their applicability. Further, \cite{ruiray22} offered additional proofs strengthening the theoretical framework of these inequalities. 
\textbf{Generalization.} Regarding generalization analyses of MTL, most studies focused on i.i.d. settings. For example, researchers leveraged tools such as Rademacher Complexity (RC) \cite{maurer06rade,maurer06linear,kakade12}, Vapnik-Chervonenkis (VC) dimension \cite{vcoriginal00}, and Covering Numbers (CN) \cite{baxter00,maurer2016benefit} to establish convergence rates of $O(\frac{1}{\sqrt{n}})$. Further, local Rademacher complexity (LRC)~\cite{yousefi18,watkins2023optimistic} was utilized, achieving a risk bound of $O(\frac{1}{n})$ under optimal conditions. For the setting of MTL with multi-graph dependent data, to our knowledge, \cite{wu2023macro-auc} is the only work, achieving a rate of $O(\frac{1}{\sqrt{n}})$ via fractional RC. Based on this work, we can obtain a fast rate of  $O(\frac{\log n}{n})$ via new techniques of concentration inequalities and local fractional RC. (see Appendix~\ref{section: D} for additional related work). 










\section{Preliminaries} 
\label{section: preliminaries}

\noindent \textbf{Notations.} Let boldfaced lowercase letters (e.g., $\va$) denote vectors, while uppercase letters (e.g., $\mA$) represent matrices. For a matrix $\mB$, denote the $i$-th row by $\vb_i$, the $j$-th column by $\vb^j$, and the element at position $(i,j)$ as $b_{ij}$. Similarly, for a vector $\vc$, $\vc_i$ denotes its $i$-th component, and the $p$-norm is represented as $\|\cdot\|_p$. $[M]$ defines the set $\{1,\dots,M\}$, with $|\cdot|$ indicating the size or cardinality of a set. 
$\Pi_N$ denotes a set satisfying $\Pi_N \coloneq \{(p_1, \dots, p_N): \sum_{i \in [N]} p_i = 1 \text{~and~} p_i \geq 0, \forall i\}$. 
For a set $A = \{a,b,c\}$, denote $\vx_A \coloneq (\vx_a,\vx_b,\vx_c)$, and thus $\vx_{[N]} \coloneq (\vx_1,\vx_2,\dots,\vx_N)$. 
The notation $\vx^{\backslash{i}}_{[N]} \coloneq (\vx_1,\dots,\vx_{i-1},\vx_{i+1},\dots,\vx_N)$ represents the vector that omits the $i$-th element, with a similar format applied to the matrix $\mX_A$ and $\mX^{\backslash{i}}_{A}$. 
Denote $\pP$ and $\eE$ as probability and expectation, respectively, such as $\pP(x \geq \epsilon) = \frac{1}{2}$ and $\eE(x)$. 
$f_g = f \circ g$ represents a composite of two functions. The variance $\mathrm{var}(f) = \eE [f(\vx)^2] - (\eE [f(\vx)])^2$ for a function $f$ over a random variable $\vx$.  

\subsection{Dependency Graph}
\label{sec:dependency_graph}
To characterize the dependency structure of random variables, we introduce some background on graph theory and dependency graph~\cite{jason04}, which helps subsequent theoretical analyses.
    \begin{definition}[Fractional independent vertex cover, and fractional chromatic number~\cite{zhang2022generalization}]
    \label{def:fractional vertex cover appendix}
        Given a graph $G = (V,E)$, where $V$ and $E$ represent the set of vertices and edges of the graph $G$, respectively. Then, there exists a set $\mathcal{D}_{G} = \{ (I_j, \omega_j) \}_{j \in [J]}$, where $I_j \subseteq V$ and $\omega_j \in (0,1]$, which is a fractional independent vertex cover of $G$, if satisfying the following: 
        \begin{enumerate}[(1)]
        \setlength\itemsep{-0.2pt}
        \vspace{-0.1in}
            \item each vertex is painted thoroughly, ensuring there are no overlaps or omissions, i.e., $\forall v \in V, \ \sum_{j: v \in I_j} \omega_j = 1$;
            \item every \( I_j \) is an independent set of vertices, which means that no two vertices in \( I_j \) are connected.
        \end{enumerate}
        \vspace{-0.1in}
        Define the fractional chromatic number $\chi_{f}(G) = \min_{\mathcal{D}_{G}} \sum_j \omega_j$, which is the minimum value among all fractional independent vertex covers of $G$.
    \end{definition}

    \begin{definition}[Dependency graph~\cite{ralaivola2015entropy}] \label{def: dependency graph}
    A series of random variables $\mX = (\vx_i)_{i=1}^N$ over $\mathcal{X}$, 
    can be associated with a corresponding dependency graph $G = (V, E)$ that illustrates the dependencies between the variables. Then, the graph $G$ satisfies the following:
        \begin{enumerate}[(1)]
        \setlength\itemsep{-0.2pt}
        \vspace{-0.1in}
            \item $V = [N]$;
            \item an edge \((j, k) \in E\) exists if and only if the random variables \(\vx_j\) and \(\vx_k\) are dependent.
        \end{enumerate}
        \vspace{-0.1in}
    \end{definition}
The above definition clarifies the relationship between the dependency graph and its associated random variables. 
This method of modeling the dependence structure of variables proves to be effective for concentration inequalities (see related work in Appendix \ref{section: D}).

Then, we define the following to characterize the property of functions of graph-dependent variables.
\begin{definition}[Fractionally colorable function~\cite{ralaivola2015entropy}] \label{def: franctionally colorable single}
    Given random variables \(\mX = (\vx_1, \ldots, \vx_N) \in \mathcal{X}^N\) with its dependency graph $G = (V, E)$.
    A function $f: \mathcal{X}^N \rightarrow \sR$ is a fractionally colorable function w.r.t. the graph $G$ if there exists a decomposition $\mathcal{D}_{G}(f) = \{ (f_{j}, I_{j}, \omega_{j}) \}_{j \in [J]}$,  satisfying: 
    \begin{enumerate}[(1)]
    \setlength\itemsep{-0.2pt}
    \vspace{-0.1in}
        \item the set \(\{ (I_{j}, \omega_{j}) \}_{j \in [J]}\) constitutes a fractional independent vertex cover of the graph \(G\);
        \item the function $f$ can be decomposed as \(f(\mX) = \sum_{j \in [J]} \omega_j f_j (\vx_{I_j})\) ,
            where $f_j: \mathcal{X}^{|I_j|} \rightarrow \sR$, \( \forall j \in [J]\).
    \end{enumerate}
\end{definition}
By Definition~\ref{def: franctionally colorable single}, we can decompose the function of single-graph dependent variables into a weighted sum of functions of independent variables, facilitating theoretical analyses. 
Further, we extend it to the function of multi-graph dependent variables, aiding in subsequent analyses of MTL with multi-graph dependent data.
\begin{definition}[Multi-fractionally sub-additive function]
\label{def: franctionally colorable multi}
    Given $m$ random variables $\mX = (\mX_1, \mX_2, \dots, \mX_K)$ with $K$ blocks, where for each $k \in [K]$, random variables $\mX_k$ is associated with a dependency graph \(G_k = ([m_k], E_k)\), and $\sum_{k \in [K]} m_k = m$. A function $f: \mathcal{X}^{m} \rightarrow \sR$ is \emph{multi-fractionally sub-additive} w.r.t. $\{G_k\}_{k=1}^K$ if it can be expressed as $f(\mX) = \sum_{k \in [K]} f_k(\mX_k)$, where each $f_k: \mathcal{X}^{m_k} \rightarrow \sR$ is fractionally colorable w.r.t. $G_k$ with a decomposition $\mathcal{D}_{G_k}(f_k) = \{ (f_{kj}, I_{kj}, \omega_{kj}) \}_{j \in [J_k]}$, 
    and each $f_{kj}$ is sub-additive (see the detailed definition 
    in Appendix~\ref{sec_app:material_sub_additive}).
\end{definition}



\subsection{Problem Setting}

Here, we introduce the learning setup of \textbf{Multi-Task Learning with Multi-Graph Dependent data (MTL-MGD)}.
Given a training dataset \( S = \{(\vx, y)\}_{i=1}^{m} \) that is organized into \( K \) 
blocks (or tasks), such that \( S = (S_1, \dots, S_K) \). Each block \( S_k = \{(\vx_{ki}, y_{ki})\}_{i=1}^{m_k} \) is drawn from a distribution \( D_k \) (for \( k \in [K] \)) over the domain \( \mathcal{X} \times \mathcal{Y} \) and is associated with a dependency graph \( G_k \), 
with \( \sum_{k \in [K]} m_k = m \). 
Let a hypothesis be ${h} \coloneq ({h}_1, \dots, {h}_K)$,
where each individual mapping \( {h}_k: \mathcal{X} \rightarrow \widetilde{\mathcal{Y}} \) corresponds to a particular task \( k \in [K] \).
Denote $\mathcal{H} \coloneq \left \{ h \right \}$ as the hypothesis space. For each $k \in [K]$, $\mathcal{H}_k \coloneq \left \{{h}_k ~|~ {h}_k: \mathcal{X} \rightarrow \widetilde{\mathcal{Y}} \right \}$, 
and define a loss function as $L: \mathcal{X} \times \mathcal{Y} \times \mathcal{H}_k \rightarrow \sR_+$. For each ${h} \in \mathcal{H}$, define its empirical risk $\widehat{R}_{S}(h)$ on dataset $S$ and expected risk $R(h)$, with their clear and simple symbols, as follows, respectively: 
    \begin{align} \label{def: initial risk}
        P_m (L_h) \defeq \widehat{R}_{S}(h) = \frac{1}{K} \sum_{k=1}^K \frac{1}{m_k} \sum_{i=1}^{m_k} L(\vx_{ki}, y_{ki}, h_k) , \quad  P (L_h) \defeq R(h) = \eE_{S} \left[ \widehat{R}_{S}(h) \right] .
    \end{align}
Similarly, for a function $g = (g_1,\dots, g_K)$, where $g_k: \mathcal{X} \rightarrow \sR, \forall k \in [K]$, 
we can define $P_m (g)$ and $P (g)$, simplified as $P_m g$ and $P g$, as follows, respectively:
\begin{align} \label{def: pmg} 
    P_m g \defeq P_m (g)  = \frac{1}{K} \sum_{k \in [K]} \sum_{j \in J_k} \frac{\omega_{kj}}{m_k} \sum_{i \in I_{kj}} g_k(\vx_i), \quad 
    P g \defeq P (g)  = \eE [P_m (g)]. 
\end{align}
The objective of MTL-MGD is to learn a hypothesis $h$ that minimizes the expected risk $P (L_h)$. 

For the distribution $D = (D_1, \dots, D_K)$, assume that there exists an optimal hypothesis $h^*=(h_1^*,\dots,h_K^*) \in \mathcal{H}$, satisfying $P (L_{h^*}) = \inf_{h \in \mathcal{H}} P (L_h)$.
In addition, in this paper, we consider the practical learning algorithm based on the Empirical Risk Minimization (ERM) rule, as follows:
\begin{align}
    \mathcal{A}: \ \hat{h} = \argmin_{h \in \mathcal{H}} P_m(L_h) .
\end{align}

\section{Main Results} \label{section: main results}


In this section, we first provide the risk bounds for MTL-MGD via a new analytical framework of local fractional RC, and then the proof techniques of new concentration inequalities. (See Figure~\ref{fig:theorem proof} in Appendix~\ref{sec-app:big_picture} for a proof structure of the main results).

\subsection{Risk Bounds for MTL-MGD}
\label{sec-mtl:risk-bound-for-mtl}
First, we give the following function class considered within this subsection:
\begin{align}
\label{eq:function_class_f}
    \mathcal{F} = \{ f: (f_1,f_2,\dots,f_K)~|~f_k: \mathcal{X} \rightarrow \sR, \forall k \in [K] \}.
\end{align}
Then, we give the following mild assumption about the loss function.
\begin{assumption}[For loss function] 
\label{thm:assump2}
Suppose a loss function satisfies the following conditions:
\begin{enumerate}[(1)]
\setlength\itemsep{-0.2pt}
    \item $L$ is $\mu $-Lipschitz continuous, i.e., $\forall k \in [K], i \in [m_k]$, $|L(\vx_{ki},y_{ki},h_k')-L(\vx_{ki},y_{ki},h_k'')| \leq \mu |h_k'(\vx_{ki})-h_k''(\vx_{ki})|$, 
    where $h_k', ~h_k'' \in \mathcal{H}_k$; 
    \item There is a constant $B > 1$ such that $\forall h \in \mathcal{H}, P(h-h^*)^2 \leq BP(L_h-L_{h^*})$.~\footnote{Note that this condition is usually met from a uniform convexity condition of $L$. In later applications of Macro-AUC and experiments, we assume $L \in [0, B]$, making this condition follow, which is easily met in practice. }
\end{enumerate} 
\end{assumption}

Then, we give a new definition of local fractional Rademacher complexity of the function class $\mathcal{F}$.
\begin{definition}[Local fractional Rademacher complexity (LFRC)]
\label{def: LFRC} 
Assume $\mathcal{F}$ is defined in Eq.~\eqref{eq:function_class_f}.~\footnote{Note that the function class can cover both the widely-used hypothesis space and loss space.} 
For a fixed $r$, the LFRC of $\mathcal{F}$ can be defined as
    \begin{align} \label{eq-def:LFRC-def}
        \mathcal{R} (\mathcal{F},r) = \eE _{S \sim D^m_{[K]}} [ \hat{\mathcal{R}} (\mathcal{F},r) ],
    \end{align}
where $S \sim D^m_{[K]}$ denotes $S_1 \sim D^{m_1}_1,S_2 \sim D^{m_2}_2,...,S_K \sim D^{m_K}_K$ for simplicity, and the empirical LFRC $\hat{\mathcal{R}} (\mathcal{F},r)$ can be defined as 
\begin{align} \label{eq: empirical LFRC}
      \hat{\mathcal{R}} (\mathcal{F},r)  
     =  \frac{1}{K} \eE _\zeta \left [ \sup _{f \in \mathcal{F},\mathrm{var}(f_k) \leq r, k \in [K]}  \sum_{k \in [K]} \frac{1}{m_k} \sum_{j \in [J_k]} \omega_{kj}  \sum_{i \in I_{kj}} \zeta_l f_k(\vx_i)\right ],
\end{align} 
where $\zeta_1,\zeta_2,...,\zeta_m$ is a sequence of independent Rademacher variables: $\pP (\zeta_l = 1) = \pP (\zeta_l = -1) = \frac{1}{2}$. 
\end{definition}
Notably, this definition is different from the classic i.i.d. one~\cite{Bartlett_2005}, as the $\omega_{kj}$ can handle (multi-)graph dependent random variables. Further, it is distinct from the fractional Rademacher Complexity (RC)~\cite{wu2023macro-auc} with multi-graph dependent variables, by incorporating additional variance information.

Then, based on LFRC and a new Talagrand-type inequality (see next subsection for details), we can derive the following base bound of Multi-Graph Dependent data (MGD) with small variance.
\begin{theorem}[The base bound of MGD with small variance, \textbf{analog of Thm. 2.1 in~\cite{Bartlett_2005}}, proof in Appendix~\ref{pro:core2.1_proof}] \label{thm:the core 2.1}
Assume the function class $\mathcal{F}$ is defined in Eq.~\eqref{eq:function_class_f}. 
Assume that there is some $r>0$ for every $f \in \mathcal{F}$, $\mathrm{var}(f_k) \leq r$, $\max_{k \in {K}} \sup_{\vx} |f_k(\vx)| \leq 1$.
Then for every $t > 0$, with probability at least $1-e^{-t}$, 
\begin{align*}
     \sup_{f \in \mathcal{F}} (P f - P_m f) 
     \leq \inf_{\alpha > 0} \left(2(1 + \alpha) \mathcal{R} (\mathcal{F} , r) + \sqrt {\frac{2crt}{K}} + (\frac{2}{3} + \frac{1}{\alpha})\frac{ct}{K} \right) ,
\end{align*}
where $c=\frac{5^2}{4^2} \sum_k \frac{\chi_f(G_k)}{m_k}$. Moreover, the same results hold for the quantity $\sup _{f \in \mathcal{F}} (P _m f - P f)$. For every $t> 0 $, with probability at least $1-e^{-t}$, 
\begin{align*}
    \sup _{f \in \mathcal{F}} (P _m f - P f) 
    \leq \inf_{\alpha > 0} \left(2(1 + \alpha) \mathcal{R} (\mathcal{F} , r) + \sqrt {\frac{2crt}{K}} + (\frac{2}{3} + \frac{1}{\alpha})\frac{ct}{K} \right) .
\end{align*}
\end{theorem}


\begin{remark}
    This theorem shows the maximum difference between the empirical and the expected value, which is associated with LFRC and the small variance of the class. 
    This theorem is the main basis of the following theoretical results. 
    Further, we can analyze the tightness of our results: when $K = 1$, our result has one more constant $\frac{5^2}{4^2}$, which cannot cover the i.i.d. case~\cite{Bartlett_2005}. But if we use Theorem \ref{thm:bennett_inequality_refined} (see Appendix \ref{sec-app:a_special_bennett_inequality}), our result can cover i.i.d. case. 
\end{remark}

In fact, the considered function class in Theorem \ref{thm:the core 2.1} lacks ``true locality'' due to the arbitrariness of the parameter $r$, restricting its practical applicability.
To address this issue, we introduce the concept of sub-root function (see Appendix \ref{sec_app:sub_root_func}), which can derive an improved and practically useful bound.
\begin{theorem}[An improved bound of MGD with sub-root function, proof in Appendix \ref{pro:theorem3.3_proof}] \label{thm: theorem 3.3 sub-root}
    



Assume that there are some functional $T : \mathcal{F}_k \rightarrow \sR^+$ and some constant $B$, for all $f \in \mathcal{F}, T(f_k) \in [\mathrm{var}(f_k), B\eE f_k]$. 
If a sub-root function $\Phi$ and its fixed point $r^*$
satisfy:~\footnote{Notably, here the content in $\{\}$ of $\mathcal{R}\{\}$ means adding some conditions to the function class (e.g., $T(f_k) \leq r$).}
\begin{align} \label{eq:def-sub-root}
    \forall r \geq r^*, ~
    \Phi (r) \geq B \mathcal{R} \{ f \in \mathcal{F}, T(f_k) \leq r \}.
\end{align}
Then for each $f\in \mathcal{F}$, $M >1$ and $t > 0$, with probability at least $1 - e^{-t}$, the following holds:
 \begin{align} \label{eq: thm sub-root 3.3_1}
    P f \leq \frac{M}{M-1} P _m f + \frac{c_1 M}{B} r^* + (c_2 BM +22 ) \frac{ct}{K},  
\end{align}
where $c_1 = 704$, $c_2 = 26$, and $c = \frac{5^2}{4^2} \sum_{k \in K}\frac{\chi_f(G_k)}{m_k} $. 
\end{theorem}
\begin{remark}
    Subsequent analyses of the fixed point facilitate practical applications of inequalities.
\end{remark}
Based on Theorem \ref{thm: theorem 3.3 sub-root}, we can analyze the excess risk bound of the algorithm $\mathcal{A}$ for MTL-MGD. 
\begin{corollary}[An excess risk bound of $\mathcal{A}$ for MTL-MGD, proof in Appendix \ref{section: A}] \label{thm : Lipschitz bound loss space}
Assume the loss function satisfies Assumption \ref{thm:assump2}. 
There exists a sub-root function $\Phi$ and its fixed point $r^*$, satisfying
        \begin{align*}
           \forall r \geq r^*, ~  \Phi (r) \geq B \mu \mathcal{R} \{ h \in \mathcal{H}, \mu^2 \eE (h_k - h_k^*)^2 \leq r \}.
        \end{align*}
    Then for every $t > 0$ and $ r \geq \Phi(r)$, with probability at least $1 - e^{-t}$, 
    \begin{align} \label{eq: Plf excess risk bound} 
        P (L_ {\hat{h}} - L_{h^*}) \leq \frac{c_1 }{B} r + (c_2 B  + 22) \frac{ct}{K},  
    \end{align}
    where $c_1 = 704$, $c_2 = 26$ and $c = \frac{5^2}{4^2} \sum_{k \in K}\frac{\chi_f(G_k)}{m_k}$. 
\end{corollary}

\begin{remark}
    From the corollary, we can observe that the rate of the risk bound is intrinsically linked to the order of $r$. 
\end{remark}

For simplicity, here we consider the kernel hypothesis space as an example of $\mathcal{H}$. (See the linear one in Proposition \ref{thm:linear upper bound} and Corollary \ref{thm: loss bound computing2} in Appendix \ref{sec-app:supplemental-risk-bounds}, and discussions about neural networks in~Appendix~\ref{section: discussion}).
First, we can have an upper bound of the LFRC of the kernel hypothesis space (see Proposition~\ref{thm:kernel upper bound} in Appendix~\ref{sec-app:supplemental-risk-bounds}).
Then, we can obtain an excess risk bound of the kernel-based algorithm $\mathcal{A}$.
\begin{corollary}[An excess risk bound of kernel-based learning algorithm $\mathcal{A}$, proof in Appendix \ref{pro:corollary2.1_proof}]\label{thm: loss bound computing}  
Suppose a loss function $L$ satisfies Assumption \ref{thm:assump2}.
Assume $\sup _{\vx \in \mathcal{X}} \kappa(\vx,\vx) < \Lambda$ and for every $k \in [K]$, $f_k = \theta_k^T \phi(\vx_k)$, where a weight vector satisfies $\|\theta_k\|_2 \leq \widetilde{M}$, and $\phi: \mathcal{X} \rightarrow \mathbb{H}$. 
For all $t > 0$, with probability at least $1 - e^{-t}$, we have \footnote{$a$ $\lesssim$ $b$ means $a$ $\leq$ $Cb$, where $C$ is a constant.}
\begin{align} 
    P (L_{\hat{h}} - L_{h^*}) \lesssim r^* + \frac{ t}{K} \sum_{k \in [K]} \frac{\chi_f(G_k)}{m_k},
\end{align}
where
\begin{align} \label{eq: loss bound computing eq}
    r^* \leq \sum _{k \in [K]} \min _{d_k \geq 0} \left( \frac{d_k \chi_f(G_k)}{K m_k} + \widetilde{M} \sqrt{\frac{\chi_f(G_k)}{K m_k} \sum _{l >d_k} \lambda_{kl}} \right).
\end{align}
\end{corollary}
\begin{remark}
    From this corollary, we discuss the order of $r^*$ w.r.t. $n$ (and $K$). For simplicity, assume $m_1 = \dots = m_K = n$, and denote $d_k^* = \argmin_{d_k \geq 0} \frac{d_k \chi_f(G_k)}{K m_k} + \widetilde{M} \sqrt{\frac{\chi_f(G_k)}{K m_k} \sum _{l >d_k} \lambda_{kl}}$. 
\begin{enumerate}[(1)]
\setlength\itemsep{-0.4pt}
    \item When for each $k \in [K], d_k^* = 0$, $r^* \leq \sum_k \widetilde{M} \sqrt{\frac{\chi_f(G_k)}{K n} \mathrm{trace}(\kappa_k)}$, we can obtain $r^* = O(\frac{1}{\sqrt{n}})$ (or $O\left(\sqrt{\frac{1}{n K} (\sum_k \chi_f(G_k))(\sum_k \mathrm{trace}(\kappa_k))}\right)$).    
    Specifically, in this case, $r^* \lesssim \sqrt{\frac{1}{n}} \cdot \sqrt{(\sum_k \frac{\chi_f(G_k)}{K}) (\sum_k \mathrm{trace} (\kappa_k))} $ (see details in Proposition~\ref{pro: kernel proof1}).
    
    \item When for some $k$,  $d_k^* \neq 0$, and the rank of kernel matrix $\mathrm{Rank}(\kappa_k) < \infty$, we can obtain $r^* = O(\frac{1}{n})$ (or $O\left(  \frac{1}{n K} \sum_k \mathrm{Rank}(\kappa_k) \chi_f(G_k)
 \right)$). Specifically, 
    we can choose $d_k' < \infty$ and $\sum_{l >d_k'} \lambda_{kl} = 0$ (i.e., $d_k' \geq \mathrm{Rank}(\kappa_k)$), thus $r^* \leq \sum_k \frac{\mathrm{Rank}(\kappa_k) \chi_f(G_k)}{K n}$.
    There are many kernels of finite rank, such as linear kernel, and we can take $d_k' = a$, where $a$ is a constant.
    \item When for every $k \in [K]$, ${\{\lambda_{kl}\}}_{l = 1}^{\infty} $ has the property of exponential decay, i.e., $\sum_{l >d_k'} \lambda_{kl} = O(e^{-d_k'})$ (e.g., the Gaussian Kernel), by setting the truncation threshold at $d_k' = \log n$, we can obtain $r^* = O(\frac{\log n}{n})$ or $O\left( \max\{\sum_k \frac{\log n \chi_f(G_k)}{n K} , \sum_k \sqrt{\frac{\chi_f(G_k)}{n^2 K}}\}\right)$.
    Specifically, in this case, $r^* \lesssim \sum_k \frac{\log n \chi_f(G_k)}{K n} + \sum_k \sqrt{\frac{\chi_f(G_k)}{n K} e^{-\log n} } \lesssim \sum_k \frac{\log n \chi_f(G_k)}{n K}+ \sum_k \sqrt{\frac{\chi_f(G_k)}{n^2 K}}$. 
    If $K \leq \log n \chi_f(G_k)$, then $r^* =O\left( \sum_k \frac{\log n \chi_f(G_k)}{n K} \right) $, else $r^* = O\left(  \sum_k \sqrt{\frac{\chi_f(G_k)}{n^2 K}}
  \right)$.
    
\end{enumerate}
\end{remark}

\subsection{Proof Techniques - New Concentration Inequalities of Multi-graph Dependent Variables}
\label{sec-con:bennett-inequality}
Here we give the proof techniques of the aforementioned risk bounds of MTL-MGD, mainly including new Bennett-type concentration inequalities for \textbf{Multi-Graph Dependent Variables (MGDV)} and Talagrand-type ones for the empirical process. For simple discussions, we give the following setup.

\textbf{Setup.} Consider $m$ random variables $\mX = (\mX_1, \mX_2, \dots, \mX_K)$ with $K$ blocks, where for each $k \in [K]$, random variables $\mX_k$ is associated with a dependency graph \(G_k = ([m_k], E_k)\), and $\sum_{k \in [K]} m_k = m$. Besides, a function $f: \mathcal{X}^{m} \rightarrow \sR$ is multi-fractionally sub-additive w.r.t. $\{G_k\}_{k=1}^K$ with the corresponding decomposition $\mathcal{D}_{G_k}(f_k) = \{ (f_{kj}, I_{kj}, \omega_{kj}) \}_{j \in [J_k]}$ for each $k \in [K]$. 
For simplified writing, we provide the following expression for each $k \in [K], \ j \in [J_k]$: 
\begin{align} \label{def :Z def}
    Z = f(\mX), \quad Z_k = f_k(\mX_k),  \quad
    Z_{kj} = f_{kj} (\vx_{I_{kj}}),  \quad Z_{kj}^{\backslash{i}} = f_{kj} (\vx_{I_{kj}^{\backslash{\{i\}}}}). 
\end{align}
Besides, the following functions are provided for subsequent analyses and discussions in this paper:
\begin{align} \label{eq:psi_function}
     G(\lambda) = \log \eE [\exp(\lambda(Z - E[Z]))], 
    \psi(x) = \exp(-x) + x - 1, 
     \varphi(x) = (1 + x) \log(1 + x) - x. 
\end{align}

Then, following~\cite{ralaivola2015entropy}, we introduce the following assumption about additional properties of the function of MGDV, for subsequent analyses. 
    \begin{assumption}
    \label{thm:assump1}
        Suppose $f$ is multi-fractionally sub-additive w.r.t. $\{G_k\}_{k=1}^K$, 
        then for every $f_k$ has a decomposition $\mathcal{D}_{G_k}(f_k) = \{ (f_{kj}, I_{kj}, \omega_{kj}) \}_{j \in [J_k]}$. 
        Besides, assume every $k \in [K], ~ j \in [J_k]$ satisfy the following conditions: 
        \begin{enumerate}[(1)]
        \setlength\itemsep{-0.4pt}
            \item for every $\vx_{I_{kji}}$, where $i \in I_{kj}$, 
            there exists a $\sigma(\vx_{I_{kj}})$- measurable random variable $Y_{kji}$ associated with it. Moreover, 
            $Y_{kji}$ satisfies: $\pP (Y_{kji} \leq Z_{kj} - Z_{kj}^{\backslash{i}} \leq 1) = 1$, and $\pP \left( \eE_{kji}[Y_{kji}] \geq 0 \right) = 1,$
            where \(\eE_{kji}\) represents the expectation relative to the \(\sigma\)-algebra formed by \((\vx_{I_{kj}^{\backslash{\{i\}}}})\);
            \item for every $\vx_{I_{kj}}$, there exists constraint values $b_{kj},~\sigma_{kj}^2 \in \sR$,  satisfying $\pP (Y_{kji} \leq b_{kj}) = 1$, and $\pP \left( \sigma_{kj}^2 \geq \sum_{i \in I_{kj}} \eE_{kji}[Y_{kji}^2] \right) = 1;$
            \item let $v_{kj} \defeq (1 + b_{kj}) \eE[Z_{kj}] + \sigma_{kj}^2$, and $v_{kj} \in \sR$.
        \end{enumerate}
    \end{assumption}
This assumption plays a role in the proof of subsequent new concentration inequalities. 
Building on prior work~\cite{ralaivola2015entropy}, we propose a new Bennett-type inequality tailored for analyses of MGDV. This new inequality both extends the existing framework and seamlessly incorporates scenarios involving single-graph dependence, broadening its applicability.
    \begin{theorem}[A new Bennett-type inequality for MGDV, proof in Appendix \ref{pro:bennett_inequ}]
    \label{thm:bennett_inequality} 
    Assume $Z$ is defined as Eq.\eqref{def :Z def}, and can be described as $Z = \sum_{k \in [K]} \sum_{j \in [J_k]} \omega_{kj} Z_{kj}$.  
    Suppose Assumption~\ref{thm:assump1} holds, and for each $k \in [K]$, $j \in [J_k]$, $b_{kj} = b$. Then, 
        \begin{enumerate}[(1)]
            \item for every $t > 0$,
                \begin{align}
                \label{eq:thm_bennet_1}
                    \pP (Z \geq \eE[Z]  + t) \leq \exp \left( -\frac{v}{W} \varphi \left( \frac{tW}{Uv}\right) \right) 
                    \leq \exp  \left( - \frac{v}{ \sum_{k \in [K]} \chi_{f} (G_k)} \varphi \left( \frac{4t}{5v}\right) \right) ,
                \end{align}
            where $\varphi$ is defined in Eq.\eqref{eq:psi_function}, $v = (1 + b) \eE [Z] + \sigma^2$, $\sigma^2 = \sum_k \sum_j \omega_{kj} \sigma_{kj}^2$, $W = \sum_{k \in [K]} \chi_f(G_k) = \sum_{k \in [K]} \omega_k$, and $U = \sum_{k \in [K]} U_k$, $U_k = \sum_{j \in J_k} \max (1, v_{kj}^{\frac{1}{2}} W^{\frac{1}{2}} v^{-\frac{1}{2}})$; 
            \item for every $t > 0$,~\footnote{Note that, Eq.~\eqref{eq:thm_bennet_2} is usually called the Bernstein-type inequality~\cite{concentration13initial}.}
                \begin{align}
                \label{eq:thm_bennet_2}
                    \pP (Z   \geq     \eE[Z] + \sqrt{2cvt} + \frac{2ct}{3}) \leq e^{-t} ,
                \end{align}
            where $c = \frac{5^2}{4^2} \sum_{k \in [K]} \chi_{f} (G_k)$.
        \end{enumerate}  
    \end{theorem}
\begin{remark}
    This is a more general Bennett inequality, encompassing the single-graph dependent one in~\cite{ralaivola2015entropy} when $K = 1$. 
    Especially, if $\omega_{kj} = 1$ for $k \in [K], ~j \in [J_k]$, then a special Bennett's inequality can be obtained (see detailed in Theorem \ref{thm:bennett_inequality_refined} of Appendix \ref{sec-app:a_special_bennett_inequality}). 
    Technically, Eq.~\eqref{eq:thm_bennet_2} is due to the employing of various common shrinking techniques, i.e., $\varphi(x) \geq \frac{x^2}{2+\frac{2x}{3}}$.~\footnote{Note that, using the existing literature 
    \cite{yousefi18,watkins2023optimistic}, we can only derive a constant of $\frac{2}{3}$ instead of $\frac{1}{3}$ (owing to the original literature \cite{bousquet2002bennett} does not provide detailed steps to prove this), but this does not affect the order of bounds.}
\end{remark}
Next, leveraging the above theorem, we derive a new Talagrand-type inequality for the empirical process. This inequality serves as a key tool for establishing LRC bounds, and consequently, improving the risk bounds for MTL-MGD.
\begin{theorem}[A new Talagrand-type inequality for empirical process, proof in Appendix \ref{section: A}]
    \label{thm:talagrand_inequality}

Let a function class $\mathcal{F} = \{ f = (f_1, f_2, \ldots, f_K) \}$, where each $f_k: \mathcal{X} \rightarrow \sR$, and assume that all functions $f_k$ are measurable, square-integrable, and fulfill the conditions $\mathbb{E} [f_k(\vx_{kj})] = 0$ for all $k \in [K]$ and $j \in [m_k]$. Besides, we require that $\|f_k\|_{\infty} \leq 1$, i.e., $\sup_{\vx} |f_k(\vx)| \leq 1$. 
Define $Z$ as follows:
    \begin{align*}
        Z \defeq \sup_{f \in \mathcal{F}} \sum_{k \in [K]} \sum_{j \in [J_k]} \omega_{kj}  \sum_{i \in I_{kj}} f_k(\vx_i).
    \end{align*}

Furthermore, for each $k \in [K]$ and $j \in [J_k]$, let $\sigma_{kj}$ represent a positive real value such that $\sigma_{kj}^2 \geq \sum_{i \in I_{kj}} \sup_{f \in \mathcal{F}} \eE [f^2(\vx_i)]$. 
Then, for every $t \geq 0$,

\begin{align}
    \pP (Z \geq \eE[Z] + t) \leq \exp \left( - \frac{v}{ \sum_{k \in [K]} \chi_{f} (G_k)} \varphi \left( \frac{4t}{5v}\right) \right) ,
\end{align}
        where $v = \sum_{k \in [K]} \sum_{j \in [J_k]} \omega_{kj} \sigma_{kj}^2 + 2 \eE [Z]$. 
        Also, with probability at least $1 - e^{-t}$,
        \begin{align}
            Z \leq \eE[Z] + \sqrt{2cvt} + \frac{2ct}{3} , 
        \end{align}
        where $c = \frac{5^2}{4^2} \sum_{k \in [K]} \chi_{f} (G_k)$.
    \end{theorem}
\begin{remark}
    Compared with \cite{wu2023macro-auc}, this theorem enables our risk bound to achieve a faster convergence rate; \cite{watkins2023optimistic} also proposes results similar to Theorem 2 under the i.i.d. case and obtains a faster rate.
\end{remark}

\section{Application to Macro-AUC Optimization}  
\label{section: applications}
In this section, we apply previous theoretical results to analyze the problem of Macro-AUC Optimization~\cite{21macro3,wu2023macro-auc,macro24,zhang2024generalization} (see other applications in Appendix~\ref{section: E}).

\textbf{Problem Setup.} Following~\cite{wu2023macro-auc} (Appendix~B.1.1 therein), we can transform the Macro-AUC optimization in multi-label learning into MTL-MGD.
This requires constructing a multi-task dataset \(S\) from the original multi-label dataset \(\widetilde{S} = \{ (\widetilde{\vx}_i, \widetilde{\vy}_i) \}_{i=1}^{\tilde{n}}\), where \( \widetilde{\vx}_i \in \widetilde{\mathcal{X}}\), \(\widetilde{\vy}_i \in \{ -1, +1 \}^K\), and \( \widetilde{y}_{ki} = 1 \) ( or \(-1\)) indicates the label \( k\) is relevant (or irrelevant). 
Besides, for each \(k \in [K]\), let \(\widetilde{S}_k^+ = \{ \widetilde{\vx}_p | \widetilde{y}_{pk} = 1, \forall p \in [\tilde{n}]\}\) and \(\widetilde{S}_k^- = \{ \widetilde{\vx}_p | \widetilde{y}_{pk} = -1, \forall p \in [\tilde{n}]\}\) denote the sets of positive and negative instances, respectively, and denote \(\tilde{n}_k^+ = | \widetilde{S}_k^+ |, \tilde{n}_k^- = | \widetilde{S}_k^- | \), and \(\tau_k = \frac{\min \{ \tilde{n}_k^+, \tilde{n}_k^- \}}{n}\) is the imbalance level factor for simplicity. The goal is to learn the best original multi-label hypothesis \( \tilde{h} \in \widetilde{\mathcal{H}} \) from \(\widetilde{S}\) to maximize the Macro-AUC metric \( \eE_{\widetilde{S}} \left[ \frac{1}{K} \sum_{k \in [K]} \frac{1}{\tilde{n}_k^+ \tilde{n}_k^-} \sum_{ (\widetilde{\vx}_p, \widetilde{\vx}_q) \in \widetilde{S}_k^+ \times \widetilde{S}_k^-} \tilde{L} (\widetilde{\vx}_p, \widetilde{\vx}_q, \tilde{h}_k ) \right] \), where \( \widetilde{\mathcal{H}} = \{(\tilde{h}_1,\tilde{h}_2,\dots, \tilde{h}_K) | \tilde{h}_k: \widetilde{\mathcal{X}} \rightarrow \sR, \forall k \in [K]\} \) and $\tilde{L}$ denotes a loss function.

In the transformation, each label is treated as an individual task, leading to \(S = \{ S_k \}_{k=1}^K\), with \(S_k = \{ (\vx_{kj}, y_{kj}) \}_{j=1}^{m_k}\). 
Specifically, for each label (or task) $k \in [K]$, each instance \(\vx_{kj}\) (\(j \in [m_k]\)) is formed from a positive and negative sample (i.e., \( \vx_{kj} = (\widetilde{\vx}_p, \widetilde{\vx}_q), \forall (\widetilde{\vx}_p, \widetilde{\vx}_q) \in \widetilde{S}_k^+ \times \widetilde{S}_k^- \)), and its label \(y_{kj} = 1\), and thus the total number of instances is \(m_k = \tilde{n}_k^+ \tilde{n}_k^- = \tilde{n}^2 \tau_k (1 - \tau_k)\). 
Besides, the transformed hypothesis \( h_k(\vx_{kj}) = \tilde{h}_k(\widetilde{\vx}_p) - \tilde{h}_k(\widetilde{\vx}_q)\). Then, we can get the dependency graph \(G_k\) of \(S_k\) based on prior results in bipartite ranking \cite{ralaivola2015entropy} satisfying: $\forall k \in [K], \chi _f(G_k) = \max \{|\tilde{n}_k^+|,|\tilde{n}_k^-|\} = (1- \tau _k) \tilde{n}.$
\textbf{Theoretical Results.} Based on the theoretical results in Section~\ref{sec-mtl:risk-bound-for-mtl}, we can get the excess risk bound of the kernel-based algorithm $\mathcal{A}$ for Macro-AUC optimization. 
\begin{corollary}[The excess risk bound of kernel-based $\mathcal{A}$ for Macro-AUC, proof in Appendix \ref{section: A}] \label{thm: kernel comput AUC}   
    Assume $\sup_{\vx} \kappa(\vx,\vx) < \Lambda$. Assume that for every $k$,  $\tilde{h}_k = \theta_k^T\phi(\widetilde{\vx})$, $\|\theta_k\|_2 \leq \widetilde{M}$, and the loss function $L$ satisfies Assumption \ref{thm:assump2}. 
    Then, with probability at least $1 - e^{-t}$, 
\begin{align} \label{eq:kernel eq AUC}
    P (L_{\hat{h}} - L_{h^*}) \lesssim r^* + \frac{t}{K \tilde{n}} \sum_{k \in [K]} \frac{1}{\tau_k},
\end{align}
where $r^* \leq \sum _{k \in [K]} \min _{d_k \geq 0} (\frac{1}{\tilde{n}} \cdot \frac{d_k}{K \tau_k} + \widetilde{M} \sqrt{\frac{1}{\tilde{n}} \cdot \frac{1}{K \tau_k} \sum _{l >d_k} \lambda_{kl}}).$
\end{corollary}


Similarly to the analysis of Corollary~\ref{thm: loss bound computing}, we can obtain a fast rate of $O(\frac{\log \tilde{n}}{\tilde{n}})$. 

\section{Experiments} 
\label{section: C}
Here, we conduct experiments to corroborate our theoretical results. We focus on an application problem of Macro-AUC optimization, studying two questions: 1) Is the condition of the upper bound of $r^*$ in InEq.\eqref{eq: loss bound computing eq} for Corollary~\ref{thm: loss bound computing} rational? 2) Is our obtained risk bound tighter than the prior~\cite{wu2023macro-auc}? 

\subsection{Experimental setup}



We use widely-used open benchmark multi-label datasets\footnote{http://mulan.sourceforge.net/datasets-mlc.html and http://palm.seu.edu.cn/zhangml/}, where some information is shown in Table~\ref{table: upper bound of pa}. Following~\cite{wu2023macro-auc}, for simplicity, we take the regularized learning algorithm with the linear model, F-norm regularizer, and (pairwise) hinge loss function, i.e., $\min_{\mW} \ \widehat{R}_{S}(h) + \lambda \| \mW \|_F^2$, where the hyper-parameter $\lambda$ is searched in $\{ 10^{-4},10^{-3},10^{-2},10^{-1} \}$ with $3-$fold cross validation.

\begin{table}
\caption{The (upper) excess risk bound values (mean $\pm$ std) of different datasets. The smaller bound is labeled as black.}
\label{table: upper bound of pa}
\centering
\begin{threeparttable} 
\begin{tabular}{lccccc}
    \toprule
    Dataset & Our bound &  Prior bound~\cite{wu2023macro-auc} & Sample size & Label size & Feature size  \\
    \midrule
    Emotions   & \textbf{3.075 $\pm$ 0.089}  & 20.750 $\pm$ 1.062  & 593 & 6 & 72 \\
    CAL500    & 31.449 $\pm$ 0.341 & \textbf{26.274 $\pm$ 0.185} & 502 & 174 & 68 \\
    Image    & \textbf{1.069 $\pm$ 0.012} & 62.070 $\pm$ 5.572  & 2000  & 5 & 294\\
    Scene    & \textbf{1.230 $\pm$ 0.006} & 28.991 $\pm$ 0.421    &   2407 & 6  & 294  \\
    Yeast    & \textbf{3.275 $\pm$ 0.217} & 6.728 $\pm$ 0.123  & 2417 & 14 & 103 \\
    Corel5k       & 134.207 $\pm$ 3.164 & \textbf{45.139 $\pm$ 0.486}  & 5000  & 374 & 499\\
    Rcv1subset1     & \textbf{23.595 $\pm$ 3.767}  & 47.538 $\pm$ 1.374 
     & 6000  & 101  & 944    \\
    Bibtex   & \textbf{6.678 $\pm$ 0.055} & 40.606 $\pm$ 1.307  & 7395 & 159 & 1836\\
    Delicious   & \textbf{6.647 $\pm$ 0.045} & 17.911 $\pm$ 0.245  & 16105 & 983 & 500\\
    \bottomrule
\end{tabular}
\end{threeparttable}
\end{table}

\subsection{Experimental results}

\noindent\textbf{The condition of the upper bound of $r^*$ is rational.} We calculate the $r^*$ via Eq.\eqref{eqappendix:the value of r}, where the result is summarized in Table~\ref{table: estimate_var}. From Table~\ref{table: estimate_var}, we can observe that $r^*$ is relatively small, indicating the condition of InEq.\eqref{eq: loss bound computing eq} holds with small values in practice, and thus the excess risk bound in Corollary~\ref{thm: loss bound computing} can follow with small values practically.
Besides, we observe $d^* = K$ in all these datasets.


\noindent\textbf{Our bound is tighter than the prior one~\cite{wu2023macro-auc}.}
We calculate the excess risk bound we obtained and the previous one~\cite{wu2023macro-auc}, where the results are summarized in Table~\ref{table: upper bound of pa}. From Table~\ref{table: upper bound of pa}, we can observe that our bound is tighter than the previous one~\cite{wu2023macro-auc} in most datasets, which verifies our theoretical results. In addition, we observe that for the CAL500 and Corel5k datasets, our bound is bigger than the previous one. This is because: in these datasets, our bound is of the order of $O(\frac{K}{n})$ (since $d^* = K$ experimentally in estimating $r^*$ for these datasets) while the previous one is $O(\frac{1}{\sqrt{n}})$; thus, when $K$ is relatively large and $n$ is relatively small (e.g. CAL500 and Corel5k datasets), the previous bound can be smaller than ours, which is also in agreement with our theoretical results.

\section{Discussion} \label{sec: discuss content}

\textbf{In terms of $K$ of the risk bound.} (See Table~\ref{table: full_discuss_nk} for details). For simplicity, assume that the rank of the kernel matrix $\mathrm{Rank}(\kappa_k)$ is finite. Compared with the one~\cite{wu2023macro-auc} in MTL-MGD setting, our result depends on $O(\sum_k \frac{\mathrm{Rank}(\kappa_k) \chi_f(G_k)}{ K})$, while~\cite{wu2023macro-auc} is of $O(\max\{\frac{1}{K} \sum_k \sqrt{\chi_f(G_k)}, \sqrt{\frac{1}{K} \sum_k \chi_f(G_k)} \})$, indicating our result and~\cite{wu2023macro-auc} are the same order of $O(1)$. Furthermore, compared with i.i.d. ones~\cite{yousefi18,watkins2023optimistic} in MTL, our result depends on $O(1)$(due to $\chi_f(G_k)=1$), while their results are of $O(\frac{1}{K})$, indicating our result is worse w.r.t. $K$, left as future work. (See more discussions in Appendix~\ref{section: discussion}).

\section{Conclusion}



In this paper, we propose a new generalization analysis for multi-task learning with multi-graph dependent data, obtaining a faster risk bound of $O(\frac{\log n}{n})$ than the previous $O(\frac{1}{\sqrt{n}})$. To establish this, we propose new techniques including new concentration equalities and a new analytical framework of local fractional Rademacher complexity with fast-rate techniques. We apply these results to Macro-AUC optimization problem, where experimental results verify our theoretical results.


\bibliographystyle{plain}
\bibliography{main2}

\newpage
\appendix 

\renewcommand{\contentsname}{Contents of Appendix}
\tableofcontents
\addtocontents{toc}{\protect\setcounter{tocdepth}{3}} 
\clearpage

\section{Background Knowledge} \label{section: F}
In this section, we will introduce some preliminary knowledge, including concepts related to handling dependent variables and basic graph theory.

\subsection{Additional material on sub-additive functions}
\label{sec_app:material_sub_additive}
\begin{definition}[Sub-additive functions, Definition~1.1 in~\cite{bousquet2003concentration}]
\label{def:sub-additive appendix}
    Given a function $f: \mathcal{X}^N \rightarrow \sR$, it is sub-additive when the following holds: 
    there exists $N$ functions, i.e., $ f_1,\dots, f_N$, where $f_i : \mathcal{X}^{N-1} \rightarrow \sR, \forall i \in [N]$, such that $\sum_{i=1}^N \left( f (\mX) - f_i (\mX^{\backslash{i}}) \right) \leq f (\mX)$, where the definition of  $\mX^{\backslash{i}}$ can be seen in Section \ref{section: preliminaries}, i.e., $\mX^{\backslash{i}} = (\vx_1, \dots, \vx_{i-1}, \vx_{i+1}, \dots, \vx_N)$. 
\end{definition}

\subsection{Material on sub-root function}
\label{sec_app:sub_root_func}
\begin{definition}[Sub-root function, Definition~3.1 in \cite{Bartlett_2005}] \label{lemma: sub-root def} 
A function \( \Phi : [0, \infty) \rightarrow [0, \infty) \) is said to be sub-root if it is nonnegative, nondecreasing, and for every \( r > 0 \), the mapping function \( r \mapsto \frac{\Phi(r)}{\sqrt{r}} \) is nonincreasing. We will only focus on nontrivial sub-root functions, meaning those that are not the constant function \( \Phi \equiv 0 \).    
\end{definition}

\begin{lemma}[The property of sub-root function, Lemma~3.2 in \cite{Bartlett_2005}] 
\label{lemma:sub-root pro} 
If a function \( \Psi : [0, \infty) \rightarrow [0, \infty) \) is a nontrivial sub-root function,
then it is continuous throughout the interval \( [0, \infty) \), and the equation \( \Phi(r) = r \) has a solitary positive solution $r^*$, which is called fixed point. 
Furthermore, for every positive value of \( r \), it follows that \( r \geq \Phi(r) \) $\Longleftrightarrow$ \( r^* \leq r \). 
\end{lemma}

\section{A big picture of the main results} 
\label{sec-app:big_picture}

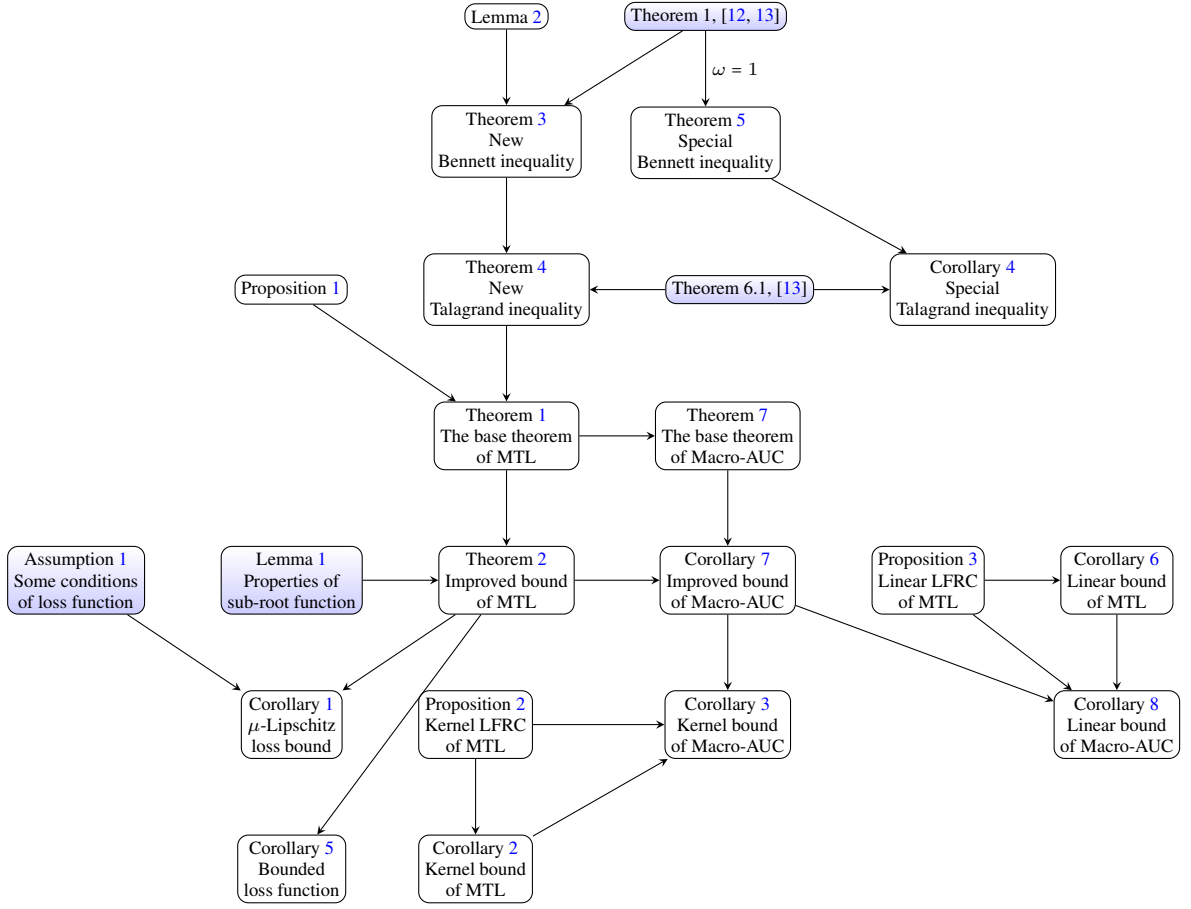
\begin{figure}[t]
        \centering
        \scriptsize
        \begin{tikzpicture}[>=stealth,
        old/.style={shape=rectangle,align=center,draw,rounded corners, top color=white, bottom color=blue!20},
        new/.style={shape=rectangle,align=center,draw,rounded corners}
        ]
    \node[new] (n1) {Lemma~\ref{pro: inequality technique}};
    \node[old] (n2) [right = of n1]
    {Theorem~1, \cite{bousquet2002bennett,bousquet2003concentration}};
    \node[new] (n3) [below = of n1]
    {Theorem \ref{thm:bennett_inequality}\\
    New \\ Bennett inequality};
    \node[new] (n4) [below = of n2] {Theorem~\ref{thm:bennett_inequality_refined}\\
    Special \\ Bennett inequality};
    \node[new] (n5) [below = of n3] {Theorem~\ref{thm:talagrand_inequality}\\
    New \\ Talagrand inequality};
    \node[old] (n6) [right = of n5] {Theorem~6.1, \cite{bousquet2003concentration}};
    \node[new] (n7) [right = of n6] {Corollary~\ref{thm:talagrand_inequality_refined}\\
    Special \\ Talagrand inequality};
    \node[new] (n8) [below = of n5] {Theorem~\ref{thm:the core 2.1}\\
    The base theorem \\ of MTL};
    \node[new] (n9) [left = of n5] {Proposition~\ref{pro: equ Rademacher}};
    \node[new] (n10) [right = of n8] {Theorem~\ref{thm: base auc}\\
    The base theorem \\ of Macro-AUC};
    \node[new] (n11) [below = of n8] {Theorem~\ref{thm: theorem 3.3 sub-root}\\
    Improved bound \\ of MTL};
    \node[old] (n13) [left = of n11] {Lemma~\ref{lemma:sub-root pro}\\
    Properties of \\ sub-root function};
    \node[new] (n14) [below = of n13] {Corollary~\ref{thm : Lipschitz bound loss space}\\
    $\mu$-Lipschitz \\ loss bound};
    \node[old] (n15) [left = of n13] {Assumption~\ref{thm:assump2}\\
    Some conditions \\ of loss function};
    \node[new] (n16) [below = of n14] {Corollary~\ref{thm: general loss bound}\\
    Bounded \\ loss function};
    \node[new] (n17) [below = of n10] {Corollary~\ref{thm: sub-root AUC}\\
    Improved bound \\ of Macro-AUC};
    \node[new] (n18) [right = of n14] {Proposition~\ref{thm:kernel upper bound}\\
    Kernel LFRC \\ of MTL};
    \node[new] (n19) [below = of n18] {Corollary~\ref{thm: loss bound computing}\\
    Kernel bound \\ of MTL
    };
    \node[new] (n20) [below = of n17] {Corollary~\ref{thm: kernel comput AUC}\\
    Kernel bound \\ of Macro-AUC};
    \node[new] (n21) [right = of n17] {Proposition~\ref{thm:linear upper bound}\\
    Linear LFRC \\ of MTL};
    \node[new] (n22) [right = of n21] {Corollary~\ref{thm: loss bound computing2}\\
    Linear bound \\ of MTL};
    \node[new] (n23) [below = of n22] {Corollary~\ref{thm: linear comput AUC}\\
    Linear bound \\ of Macro-AUC};

    \draw[->] (n1) -- (n3);
    \draw[->] (n2) -- node[right] {$\omega=1$} (n4);
    \draw[->] (n3) -- (n5);
    \draw[->] (n2) -- (n3);
    \draw[->] (n6) -- (n5);
    \draw[->] (n6) -- (n7);
    \draw[->] (n5) -- (n8);
    \draw[->] (n8) -- (n11);
    \draw[->] (n4) -- (n7);
    \draw[->] (n8) -- (n10);
    \draw[->] (n9) -- (n8);
    \draw[->] (n13) -- (n11);
    \draw[->] (n11) -- (n17);
    \draw[->] (n11) -- (n14);
    \draw[->] (n11) -- (n16);
    \draw[->] (n15) -- (n14);
    \draw[->] (n17) -- (n23);
    \draw[->] (n17) -- (n20);
    \draw[->] (n18) -- (n20);
    \draw[->] (n19) -- (n20);
    \draw[->] (n18) -- (n19);
    \draw[->] (n21) -- (n23);
    \draw[->] (n21) -- (n22);
    \draw[->] (n22) -- (n23);
    \draw[->] (n10) -- (n17);

        \end{tikzpicture}
        \caption{The proof structure diagram of the main results. The blue node denotes previous results (or assumptions) and others are our contributions.}
        \label{fig:theorem proof}
    \end{figure}

In this section, we present a diagrammatic framework in~Figure~\ref{fig:theorem proof} outlining the proofs of the main theorems and corollaries, providing a concise overview of the content in Sections \ref{section: G} $\sim$ \ref{section: E}. This will facilitate an understanding of the primary theoretical outcomes of this paper and their relationship to previous results.


\section{Additional Concentration Inequalities
}

\label{section: G}

This section serves as a supplement to Section \ref{section: main results}, discussing specific cases of the new Bennett inequality proposed in this paper, as well as some direct applications, such as the Talagrand inequality.
\subsection{A Special Bennett's Inequality}
\label{sec-app:a_special_bennett_inequality}
    \begin{theorem}[A new refined Bennett's inequality for a special case of multi-graph dependent variables, proof in Appendix \ref{pro:theorem4_proof}]
    \label{thm:bennett_inequality_refined}
    $Z = \sum_{k \in [K]} \sum_{j \in [J_k]} \omega_{kj}Z_{kj} $ (see details in Eq. \eqref{def :Z def}, Section \ref{section: main results}).  
    Suppose Assumption \ref{thm:assump1} holds, and for every $k \in [K], j \in [J_k]$,  $b_{kj}=b, ~\omega_{kj}=1$. Then 

        
        \begin{enumerate}[(1)]
            \item for every $t > 0$,
                \begin{align}
                \label{eq:thm_bennet_1_new}
                    \pP (Z \geq \eE[Z] + t) \leq \exp \left( - v \varphi \left( \frac{t}{v { \sum_{k \in [K]} \chi_{f} (G_k)} } \right) \right) ,
                \end{align}
            where $v = (1 + b) \eE [Z] + \sigma^2$, $W = \sum_{k \in [K]} \chi_f(G_k) = \sum_{k \in [K]} \omega_k$, and $\omega_k = |J_k|$, $U = \sum_{k \in [K]}$. Additionally,  $\varphi$ can be seen in  Eq.\eqref{eq:psi_function}, i.e., $\varphi(x) = (1+x) \log(1+x)-x$;
            \item for every $t > 0$,
                \begin{align}
                \label{eq:thm_bennet_2_new}
                    \pP (Z \geq \eE [Z] + c \sqrt{2vt} + \frac{2ct}{3}) \leq e^{-t} ,
                \end{align}
                where $c = \sum_{k \in [K]} \chi_{f} (G_k)$ (details in Appendix \ref{section: B}).  
        \end{enumerate} 

    \end{theorem}
\begin{remark}
    This theorem serves as a complement to Theorem \ref{thm:bennett_inequality} and represents a specific instance of the Bennett inequality, applicable to i.i.d random variables. A detailed analysis and discussion of its implications will follow.
\end{remark}
Discussion:
then we discuss the above inequality in two cases, and in contrast to the i.i.d. case \cite{Bartlett_2005,yousefi18,watkins2023optimistic} and the single graph \cite{ralaivola2015entropy}. 
\begin{enumerate}[(1)]
    \item in the i.i.d. case, $\chi _f(G_k) = \chi _f(G) = 1 $, we take $K = 1$ to contrast to single-graph, \\
    the result in the single-graph case is  
    \begin{align*}
       \pP (Z - \eE Z \geq t) \leq e^{-\frac{v}{\chi _f(G)} \varphi (\frac{4t}{5v})} = e^{-v \varphi (\frac{4t}{5v})}, 
    \end{align*}
    while our result is  
    \begin{align*}
        \pP (Z - \eE Z \geq t) \leq e^{-v \varphi (\frac{t}{v \sum _{k \in [K]} \chi _f(G_k)})} = e^{-v \varphi (\frac{t}{v})}.
    \end{align*} 
For a fixed $t$, we observe that the smaller $ \pP (Z - \eE Z \geq t) $ this term, the better. Since $\varphi (\frac{t}{v}) > \varphi (\frac{4t}{5v})$, $e^{-v \varphi (\frac{t}{v})} < e^{-v \varphi (\frac{4t}{5v})}$, our bound is tighter, and Bennett's inequality in prior work can be viewed as a special case. 

    \item in the graph-dependent case, $\chi_f(G) \neq 1$, we also take $K =1$ to contrast to the single-graph. In this case, $\sum _{k \in [K]} \chi _f(G_k) = \chi _f(G) \geq 2$,  \\
    the result in \cite{ralaivola2015entropy} is  
    \begin{align*}
       \pP (Z - \eE Z \geq t) \leq e^{-\frac{v}{\chi _f(G)} \varphi (\frac{4t}{5v})}, 
    \end{align*}
    while our result is  
    \begin{align*}
        \pP (Z - \eE Z \geq t) \leq e^{-v \varphi (\frac{t}{v \sum _{k \in [K]} \chi _f(G_k)})} = e^{-v \varphi (\frac{t}{v \chi _f(G)})}.
    \end{align*}

Denote $\clubsuit _1 = -\frac{v}{\chi _f(G)} \varphi (\frac{4t}{5v}) $, and $ \clubsuit _2 = -v \varphi (\frac{t}{v \chi _f(G)}) $, $c_1 = \frac{1}{\chi _f(G)}$, $c_2 = \frac{4}{5}$, $x = \frac{t}{v}$, then
\begin{align*}
    \clubsuit_1 - \clubsuit _2 
    = &  v \varphi (c_1 x) - c_1 v \varphi (c_2 x) 
    =  v (\varphi (c_1 x) - c_1 \varphi(c_2 x) ) \\ 
    = & (1 +c_1 x)\log (1+c_1 x)   -  c_1(1 +c_2x) \log (1+ c_2 x) - (c_1 - c_1 c_2)x.
\end{align*}
We can notice that as $\chi _f(G)$ increases, $\clubsuit _1 - \clubsuit_2 $ becomes smaller and smaller, i.e., $\clubsuit _1 - \clubsuit _2 \approx 0 $. Thus our result in Theorem \ref{thm:bennett_inequality_refined} is equivalent to the result in \cite{ralaivola2015entropy}. 

\end{enumerate}

Similar to Theorem \ref{thm:talagrand_inequality}, we can derive the special Talagrand-type inequality from Theorem \ref{thm:bennett_inequality_refined}.

\subsection{A Special Talagrand-type Inequality}
\label{sec-app:special_talagrand_inequality}
\begin{corollary} [A new refined Talagrand-type inequality for empirical process with a special case of  multi-graph dependent variables, proof in Appendix \ref{section: A}]
\label{thm:talagrand_inequality_refined}
    Denote $\mX$ as some random variables, which are divided in the same way as the Section \ref{section: preliminaries}, i.e., $\mX = (\mX_1,\mX_2,\dots,\mX_K)$. For every $k \in [K]$, $\mX_k = (\vx_{k1},\vx_{k2},\dots,\vx_{k m_k})$, with $m=\sum_{k \in [K]}m_k$. Assume that each $\mX_k$ is linked to a dependence graph $G_k$, where $\{ (I_{kj}, \omega_{kj}) \}_{j \in [J_k]}$ constitues a fractional independent vertex cover of $G_k$, and define $\chi_{f} (G_k) \defeq \sum_{j \in [J_k]} \omega_{kj}$.
    


Let $\mathcal{F} = \{ f, f = (f_1,\dots,f_K) \}$, where each $f_k: \mathcal{X} \rightarrow \mathcal{Y}$, and assume that all functions $f_k$ are measurable, squareintegrable, and fulfill the conditions $\eE [f_k(\vx_{kj})] = 0, ~\omega_{kj}=1$ for all $k \in [K]$ and $j \in [m_k]$. Additionally, we require that $\|f_k\|_{\infty} \leq 1$.
Define $Z$ as follows:
    \begin{align*}
        Z \defeq \sup_{f \in \mathcal{F}} \sum_{k \in [K]} \sum_{j \in [J_k]}  \sum_{i \in I_{kj}} f_k(\vx_i).
    \end{align*}


Moreover, for every $k \in [K]$, $j \in [J_k]$, donate a positive real value $\sigma_{kj}$, satisfying $\sigma_{kj}^2 \geq \sum_{i \in I_{kj}} \sup_{f \in \mathcal{F}} \eE[f^2(\vx_i)]$. Then, for every $t \geq 0$,
\begin{align}
     \pP (Z \geq \eE[Z] + t) \leq \exp \left( - v \varphi \left( \frac{t}{v W}\right) \right) ,
\end{align}
where $v = \sum_{k \in [K]} \sum_{j \in [J_k]} \sigma_{kj}^2 + 2 \eE [Z]$, $W = \sum _{k \in [K]} \chi _f(G_k) $, and the definition of other variables in Theorem \ref{thm:bennett_inequality_refined}. 
Also, with probability at least $1 - e^{-t}$,
\begin{align}
Z \leq \eE[Z] + c \sqrt{2vt} + \frac{2ct}{3} , 
\end{align}
where $c = \sum_{k \in [K]} \chi_{f} (G_k)$.

    \end{corollary}


\begin{remark}
    This result is specific to Theorem \ref{thm:talagrand_inequality} and represents an inequality directly derived from Theorem \ref{thm:bennett_inequality_refined}. It can be used to further develop the form of LRC and applies to the generalized boundary analysis of MTL, encompassing the case of i.i.d. (detail information and in \ref{section: A}). 
\end{remark}

\subsection{Supplement to Bennett Inequality}
\label{sec-app:supplement-bennett-lower}
\begin{theorem}[The lower bound of Bennett's inequality for a special case of multi-graph dependent variables] \label{thm: bennett inequality lower bound}
Given $Z = \sum_{k \in [K]} \sum_{j \in [J_k]} \omega_{kj}Z_{kj} $ (see details in Eq.\eqref{def :Z def}, Section \ref{section: main results}).  Suppose Assumption \ref{thm:assump1} holds, and for every $k \in [K], j \in [J_k]$, $b_{kj}=b, ~\omega_{kj}=1$. Then for all $t>0$, 
\begin{align}
    \label{eq:thm_bennet_lower}
    \pP (Z \leq \eE[Z] - t) \leq \exp \left( - \frac{v}{W} \varphi \left( \frac{tW}{Uv} \right) \right) \leq \exp \left( -\frac{v}{\sum_{k \in [K]} \chi_f(G_k) } \varphi \left(\frac{4t}{5v} \right) \right) ,
\end{align}
where $\varphi$ is defined as above, i.e., $\varphi(x)=(1+x) \log(1+x)-x$. Owing to the fact $x \geq 0$, $\varphi(x) \geq \frac{x^2}{2+ \frac{2x}{3}}$, we can get 
\begin{align}
    \label{eq:thm-bennett_lower2}
    \pP (Z \leq \eE [Z] - \sqrt{2cvt} - \frac{2ct}{3}) \leq e^{-t},
\end{align}
where $c = \frac{5^2}{4^2} \sum_{k \in [K]} \chi_f(G_k)$. The definition of $v,~W,~U$ can be seen in Theorem \ref{thm:bennett_inequality}.      
\end{theorem}

\begin{remark}
    This theorem supplements the Bennett inequality and can be used to analyze the lower bounds of the generalization bounds discussed later. Furthermore, this theorem can be combined with the results of Theorem \ref{thm:bennett_inequality}, then $\pP (|Z - \eE [Z]| \geq t) \leq 2 \exp ( -\frac{v}{\sum_{k \in [K]} \chi_f(G_k)} \varphi (\frac{4t}{5v} ) )$. The equivalence of the upper and lower bounds implies that the bound obtained is relatively tight. However, the result does not cover the i.i.d. case. This indicates that the result is not optimal and that there is still a gap $\frac{5}{4}$  to be bridged.
\end{remark}

\begin{proof}
    We can define some constants as Theorem \ref{thm:bennett_inequality}, then
    \begin{align*}
        \pP (Z - \eE[Z] \leq -t) = \pP (\eE[Z] - Z \geq t) = \pP (e^{\lambda (\eE[Z]-Z)} \geq e^{\lambda t}) \leqone \eE(e^{\lambda (\eE[Z]-Z)}) \cdot e^{-\lambda t},
    \end{align*}
\text{\ding{172}} is owing to the fact, i.e.,  $\forall a >0,~ \pP [x >a] \leq \frac{\eE[X]}{a}$. The result can then be obtained using the same technique as in the proof of Theorem \ref{thm:bennett_inequality}. 
\end{proof}

\section{Theoretical Results for MTL with Multi-graph Dependent Data} 
\label{section: MTL gereral bound}

In this section, we will provide some results on the generalization bounds of MTL in the case of graph-dependent case, to supplement the content of Section \ref{section: main results} in the main text.
\subsection{Supplemental definition and proposition}
\begin{definition}[The LFRC of the loss space]
    Define $\mathcal{H}=\{ h = (h_1,h_2,...,h_K) | h_k: \mathcal{X} \rightarrow \widetilde{\mathcal{Y}}, k \in [K] \}$ as the hypothesis space, $\mathcal{H}_k = \{ h_k: \mathcal{X} \rightarrow \widetilde{\mathcal{Y}} \}$, and $L : \mathcal{X} \times \mathcal{Y} \times \mathcal{H}_k \rightarrow \sR^+$, $L \in [0,M_c]$. According to Definition \ref{def: LFRC}, the empirical LFRC of loss space is defined as
    \begin{align*}
        \hat{\mathcal{R}}_S (L \circ \mathcal{H}) = \frac{1}{K} \eE _\zeta \left[ \sup _{h \in \mathcal{H}, \mathrm{var}(h_k) \leq r} \sum _{k \in [K]} \frac{1}{m_k}   \sum _{j \in J_k} \omega_{kj}  \sum _{i \in I_{kj}} \zeta_i L(x_i,y_i,h_k)   \right].
    \end{align*}        

    Furthermore, the LFRC of $L \circ \mathcal{H}$ is defined as
    \begin{align*}
        \mathcal{R}_m (L \circ \mathcal{H}) = \eE _{S \sim D^m_{[K]}} [\mathcal{R}_S(L \circ \mathcal{H})].
    \end{align*}    
\end{definition} 

\begin{proposition} \label{pro: equ Rademacher}
For every $ r > 0 $, 
\begin{align}
    \eE _{S,S'}  \sup _{f \in \mathcal{F}} \sum _{k \in [K]} \sum _{j \in [J_k]}  \frac{\omega_{kj}}{K m_k}  \sum _{i \in I_{kj}} ( f_k(\vx'_i) - f_k(\vx_i)) \leq 2 \mathcal{R} (\mathcal{F},r) .
\end{align}

\end{proposition}
we can prove this inequality due to the property of the Rademacher Complexity and the symmetry.  
\subsection{Supplemental risk bounds}
\label{sec-app:supplemental-risk-bounds}
\begin{corollary}[An excess risk bound of learning multiple tasks with graph-dependent examples, proof in Appendix~\ref{section: A}] \label{thm: general loss bound}
    Let \( \hat{h} \) and \( h^* \) denote the prediction functions, which correspond to the minimum empirical loss and the minimum expected loss, respectively.
    Assume a bounded loss function $ L \in [0, M_c] $. Assume:

    a sub-root function $\Phi$ and its fixed point $r^*$ satisfied the following: 
    \begin{align*}
        \forall r \geq r^*,~  \Phi (r) \geq M_c \mathcal{R} \{ h \in \mathcal{H}, \eE (L_{h_k} - L_{h_k^*})^2 \leq r \}.
    \end{align*}
    Then for every $t > 0$, with probability at least $1 - e^{-t}$, 
    \begin{align} \label{eq: collary 5.1 bounded loss}
        P (L_ {\hat{h}} - L_{h^*}) \leq \frac{c_1 }{M_c} r^* + (c_2 M_c  + 22) \frac{ct}{K},  
    \end{align}
    where $c_1 = 704$, $c_2 = 26$, and $c = \frac{5^2}{4^2} \sum_{k \in K}\frac{\chi_f(G_k)}{m_k} $. 
\end{corollary}

This corollary gives the risk bound of a general bounded loss function, which can be improved if some properties of the loss function and the hypothesis class are taken into account. 
\begin{proposition}[The upper bound of LFRC of the kernel hypothesis space, proof in Appendix \ref{pro:proposition1_proof}] \label{thm:kernel upper bound}
Define $\kappa: \mathcal{X} \times \mathcal{X} \rightarrow \sR$ as a Positive Definite Symmetric (PDS) kernel and let its induced reproducing kernel Hilbert space (RKHS) be $\mathbb{H}$. Assume for every $k \in [K]$, $f_k = \theta_k^T \phi(\vx_k)$, where a weight vector satisfies $\|\theta_k\|_2 \leq \widetilde{M}$, and $\phi: \mathcal{X} \rightarrow \mathbb{H}$. 
For every $r >0$, the LFRC of function class $\mathcal{F}$ satisfies 
    \begin{align*}
        \mathcal{R}\{ f \in \mathcal{F}, \eE f_k^2 \leq r \}  
        \leq \sum _{k \in [K]} \left(\frac{2 \chi_f(G_k)}{K m_k} \sum _{l=1}^{\infty} \min \{ r, \widetilde{M}^2 \lambda_{kl} \} \right)^{\frac{1}{2}},
    \end{align*}
where for every $k \in [K]$, the eigenvalues $(\lambda _{kl})_{l=1} ^{\infty}$
are arranged in a nonincreasing order, which satisfies $\kappa_k (\vx,\vx') = \sum _{l =1} ^{\infty} \lambda_{kl} \varphi_{kl}(\vx)^T \varphi _{kl}(\vx') $. 

Moreover, if for every $k \in [K]$, $\lambda_{k 1} \geq \frac{1}{m_k \widetilde{M}^2}$. 
Then for every $r \geq \frac{1}{m}$ and $m=\sum_{k \in [K]} m_k$, 
\begin{align*}
    \mathcal{R}\{ f \in \mathcal{F}, \eE f_k^2 \leq r \}  
    \geq c \sum _{k \in [K]} \left(\frac{ \chi_f(G_k)}{K m_k} \sum _{l=1}^{\infty} \min \{ r, \widetilde{M}^2 \lambda_{kl} \}\right)^{\frac{1}{2}},
\end{align*}
where $c$ is a constant. 
\end{proposition}

\begin{remark}
    This proposition indicates the upper bound of $\mathcal{R}(\mathcal{F},r)$ is tight since it matches the order of its lower bound.
\end{remark}

\begin{proposition}[The upper bound of LFRC in linear hypothesis, proof in Appendix \ref{pro:proposition3_proof}]
\label{thm:linear upper bound} Assume that $\sup _{\vx \in \mathcal{X}} \|\vx\|_2^2 \leq \bar{M}^2$, $\bar{M} >0$. The function class $\mathcal{F} = \{ f, f = (f_1,f_2,\dots,f_K), f_k = \theta_k^T \vx_k, \|\theta_k\|_2 \leq \widetilde{M} \}$. For every $ r >0$, 
    \begin{align*}
       & \mathcal{R}\{ f \in \mathcal{F}, \eE f_k^2 \leq r \} \leq \sum _{k \in [K]} \left( \frac{2 \chi_f(G_k)}{K m_k} \sum _{kl = 1} ^{\infty} \min \{ \frac{r}{\bar{M}^2} , \widetilde{M}^2 \widetilde{\lambda} ^2_{l} \} \right)^ {\frac{1}{2}}. \\
    \end{align*}
where singular values $(\widetilde{\lambda}_l)_{l=1}^{\infty}$ in a nonincreasing order, and $\Theta = \sum_{l=1}^{\infty}u_l v_l^T \widetilde{\lambda}_l$, where $\Theta$ is a weight matrix. 

Moreover, if $\widetilde{\lambda}_1^2 \geq \frac{1}{m \widetilde{M}^2}$, then for every $r \geq \frac{\bar{M}^2}{m}$, $m = \sum_{k \in [K]} m_k$,
\begin{align*}
     \mathcal{R}\{ f \in \mathcal{F}, \eE f_k^2 \leq r \}  
     \geq c \sum_k \left(  \frac{ \chi_f(G_k)}{K m_k} \sum _{l = 1} ^{\infty} \min \{ \frac{r}{\bar{M}^2} , \widetilde{M}^2 \widetilde{\lambda} ^2_{l} \} \right)^{\frac{1}{2}},    
\end{align*}
where c is a constant. 
\end{proposition}

\begin{remark}
    The above gives the upper and lower bounds for linear space.
    Furthermore, the relationship between this upper bound and the fixed point \( r^* \) is examined to facilitate a detailed analysis of the complexity associated with the generalization bound. 
\end{remark}

\begin{corollary}[An excess risk bound in linear hypothesis, proof in Appendix \ref{pro:corollary6_proof}]
\label{thm: loss bound computing2} Assume that $\sup _{\vx \in \mathcal{X}} $ $ \|\vx\|_2^2 \leq \bar{M}^2,\bar{M} > 0$, $\|\theta_k\|_2 \leq \widetilde{M}$, and loss function $L$ satisfies Assumption \ref{thm:assump2}, $C$ is a constant about $B, \mu $, and $C'$ is a constant about $\chi_f(G)$. Then for all $t > 0$,  with probability at least $1 - e^{-t}$, 
\begin{align}
    P (L_{\hat{h}} - L_{h^*}) \leq C_{B,\mu} (r^* + C'_{\chi_f(G)} \frac{t}{K}),
\end{align}
where
\begin{small}
    \begin{align} \label{eq:loss bound computing2 eq}
    r^* \leq \sum_k \min_{d \geq 0} ( \frac{d}{\bar{M}^2} \frac{\chi_f(G_k)}{K m_k} + \widetilde{M} \sqrt{ \frac{\chi_f(G_k)}{K m_k} \sum_{l > d} \widetilde{\lambda}_{l}^2 } ) ,
    \end{align}
\end{small}
where $d$ is the division of singular values of matrix $\Theta$, and $P (L_{\hat{h}} - L_{h^*})$ can be seen in Theorem \ref{thm : Lipschitz bound loss space}. 
\end{corollary}

\section{Proofs} \label{section:proof_all}
In this section, we will give detailed proofs of some major theorems and corollaries, as well as some simple proof techniques for corollaries, so that readers can understand and sort out the main results.
\subsection{Proof Sketches} \label{section: A}
This section outlines the main ideas and techniques used in the proofs of several theorems and corollaries.

(1) Proof of Theorem \ref{thm:talagrand_inequality}, we can use Theorem 6.1 in \cite{bousquet2003concentration} and Theorem \ref{thm:bennett_inequality} to get the following: 
\begin{itemize}
    \item $ Y_{kjl} \leq Z_{kj}-Z_{kj}^{\backslash \{ l \} } \leq 1$, these inequalities are owing to 
    \begin{align*}
        & \sum_{i \in I_{kj} \backslash \{l \} } f_l^{kj}(\vx_i)  = \sup_{f \in \mathcal{F}} \sum_{i \in I_{kj} \backslash \{l \}} f(\vx_i), ~
         Y_{kjl} = f_l^{kj}(\vx_l),  \\
        & Y_{kjl} \leq Z_{kj}-Z_{kj}^{\backslash \{ l \} } \leq f_{kj}^*(\vx_l) \leq 1, ~
        \eE_{kjl} [Y_{kjl}] = 0. \\ 
    \end{align*}    

    \item $\sigma_{kj}^2 \geq \sum_{l \in I_{kj}} \eE _{I_{kj}} [Y_{kjl}^2]$, because
    \begin{align*}
        \sum_{l \in I_{kj}} \eE _{I_{kj}} [Y_{kjl}^2] = \sum_{l \in I_{kj}} \eE _{I_{kj}} [f_l^{{kj}^2}(\vx_l)] \leq \sum_{l \in I_{kj}} \sup_{f \in \mathcal{F}} \eE [f^2 \vx_l)].
    \end{align*}
\end{itemize}

(2) Proof of Corollary \ref{thm:talagrand_inequality_refined}, similar to above, we can use Theorem \ref{thm:bennett_inequality_refined} and Theorem 6.1 in \cite{bousquet2003concentration} to get the results, and we can observe $b = 1$. 

(3) Proof of Corollary \ref{thm: general loss bound}, let $g = L_{\hat{h}} - L_{h^*}$, $T(g_k) = \eE g_k^2$ and we notice $g_k \in [-M_c,M_c]$, 
then $\mathrm{var}(g_k) = \eE g_k^2 - (\eE g_k)^2 \leq \eE g_k^2$, i.e., $ \mathrm{var}(g_k) \leq T(g_k) $, $T(g_k) = \eE g_k^2 \leq M_c \eE g_k$,
then we can use Theorem \ref{thm: theorem 3.3 sub-root} to $g$, and since $P_m g  \leq 0$, we can omit the term $\frac{M}{M-1} P_m g$.

(4) Proof of Corollary \ref{thm : Lipschitz bound loss space}, according to Assumption \ref{thm:assump2}, then $\eE (L_{h_k} - L_{h_k^*})^2 \leq \mu^2 \eE (h_k - h_k^*)^2$, let $T(L_{h_k} - L_{h_k^*}) = \eE (L_{h_k} - L_{h_k^*})^2 \leq B \mu ^2 \eE (L_{h_k} - L_{h_k^*}) $, and we know $\mu \mathcal{R} \{ h,\mu^2 \eE(h_k - h_k^*) ^2 \leq r \} \geq \mathcal{R} \{ L_h - L_{h^*}, \mu ^2 \eE (h_k - h_k^*)^2 \leq r \}$. And we notice $P _m(L_{\hat{h}} - L_{h^*}) \leq 0 $, then use Theorem \ref{thm: theorem 3.3 sub-root} to get the results.

(5) Proof of Theorem \ref{thm: base auc}, we can use Theorem \ref{thm:the core 2.1} and the value of $m_k, 
 \chi_f(G_k)$, i.e., $m_k = \tilde{n}^2 \tau_k(1-\tau_k)$, $\chi_f(G_k) = (1- \tau_k)\tilde{n}$, to obtain the results. 

(6) Proof of Corollary \ref{thm: sub-root AUC}, we can prove the corollary by using Theorem \ref{thm: base auc} and Theorem \ref{thm: theorem 3.3 sub-root}, the value of $m_k, \chi_f(G_k)$, that is, $m_k = \tilde{n}^2 \tau_k(1-\tau_k)$, $\chi_f(G_k) = (1- \tau_k)\tilde{n}$.

(7) Proof of Corollary \ref{thm: kernel comput AUC}, we can use Corollary \ref{thm: sub-root AUC} and Proposition \ref{thm:kernel upper bound} (the upper bound of $r^*$), \ref{thm: loss bound computing} (the risk bound in the kernel hypothesis) to get the results and use the value of $m_k, \chi_f(G_k)$, i.e., $m_k = \tilde{n}^2 \tau_k(1-\tau_k)$, $\chi_f(G_k) = (1- \tau_k)\tilde{n}$. Also, if we use one kernel matrix, we can get $r^* \leq \min _{0 \leq d \leq m} \left(\frac{1}{\tilde{n}} \sum_{k \in [K]} \frac{1}{\tau_k} d + \widetilde{M} \sqrt{\frac{1}{\tilde{n}} \sum_{k \in [K]} \frac{1}{ \tau_k} \sum _{l >d} \lambda_{l}} \right)$. 

(8) Proof of Corollary \ref{thm: linear comput AUC}, we can use Corollary \ref{thm: sub-root AUC} and Proposition \ref{thm:linear upper bound} (the upper bound of $r^*$), \ref{thm: loss bound computing2} (the risk bound in linear hypothesis), and use the value of $m_k, \chi_f(G_k)$, i.e., $m_k = \tilde{n}^2 \tau_k(1-\tau_k)$, $\chi_f(G_k) = (1- \tau_k)\tilde{n}$. 


\subsection{Proof Details} 
\label{section: B}
This section provides a detailed explanation of the proof procedures for several key theorems and lemmas, aiming to facilitate understanding of their underlying reasoning.

\subsubsection{Proof of Theorem \ref{thm:bennett_inequality}}
\label{pro:bennett_inequ}
Here we give the proof of \textbf{Theorem \ref{thm:bennett_inequality} (A new Bennett's inequality for multi-graph dependent variables)}, 
and review $Z = \sum_{k \in [K]} \sum _{j \in [J_k]} \omega_{kj} Z_{kj} $, where $Z_{kj} = f_{kj}(x_{I_{kj}})$. 
\begin{proof}
Define some constants as follows:
        \begin{align*}
            & \sigma^2 = \sum_{k \in [K]} \sum_{j \in [J_k]} \omega_{kj} \sigma_{kj}^2, \ v = (1 + b) \eE [Z] + \sigma^2, 
             c = \frac{5^2}{4^2} \sum_{k \in [K]} \chi_{f} (G_k), \ \chi_{f} (G_k) \defeq \sum_{j \in [J_k]} \omega_{kj} , \\
            & p_k = \frac{U_k}{U}, q_{kj} = \frac{\omega_{kj} \max (1, \overbrace{v_{kj}^{\frac{1}{2}} W^{\frac{1}{2}} v^{-\frac{1}{2}} } ^{\triangle})}{U_k}, 
             U = \sum_{k \in [K]} U_k, U_k = \sum_{j \in J_k} \omega_{kj} \max(1, \triangle), \\
            & W = \sum_{k \in [K]} \chi_f(G_k) = \sum_{k \in [K]} \omega_k,  
             \sigma _k^2 = \sum_{j \in I_{J_k}} \omega_{kj} \sigma _{kj}^2, \omega_k = \sum_{j \in [J_k]} \omega_{kj}, 
             v_k = \sum_{j \in [J_k]} \omega_{kj}v_{kj} = (1+b) \eE [Z_k] +\sigma _k^2. 
        \end{align*}
Then
    \begin{align*}
        \pP(Z-\eE Z \geq t) & = \pP(e^{\lambda (Z- \eE Z)} \geq e^{\lambda t}) \leqone \underbrace{  \eE(e^{\lambda (Z- \eE Z)}) }_{e^{G(\lambda )}} \cdot e^{-\lambda t}. \\
       G(\lambda )  & = \log \eE (e^{\lambda (Z - \eE Z)}) \\
        & = \log \eE (e^{\lambda \sum _{k \in [K]}p_k \frac{1}{p_k}  (Z_k - \eE _{X_k} [Z_k] )})  \\
        & = \log \eE (e^{\lambda \sum _{k \in [K]}p_k \frac{1}{p_k} \sum _{j \in [J_k] } q_{kj} \frac{\omega_{kj}}{q_{kj}}  (Z_{kj} - \eE _{X_{I_{kj}}}  [Z_{kj}] )}) \\
        & \leqtwo  \log \sum_{k \in [K]}p_k\sum_{j \in [J_k]}q_{kj} 
        \eE(e^{\frac{\lambda \omega_{kj}}{p_k q_{kj}}(Z_{kj}- \eE [Z_{kj}]) }) ,\\   
        G_{kj}(\lambda ) & = \log \eE (e^{\lambda (Z_{kj}- \eE Z_{kj})}), \\
        G(\lambda ) & \leqthree \log \sum_{k \in [K]}p_k\sum_{j \in [J_k]}q_{kj} (e^{ G_{kj} ({\frac{\lambda \omega_{kj}}{p_k q_{kj}} })}) \\
        & \leqfour \log \sum_{k \in [K]}p_k\sum_{j \in [J_k]}q_{kj} (e^{ v_{kj} \psi (-{\frac{\lambda \omega_{kj}}{p_k q_{kj}} })}).
    \end{align*}

\text{\ding{172}} is due to the Markov Inequality, i.e., $\pP [x > a] \leq \frac{\eE[x]}{a}$, and we define $e^{G(\lambda) } = \eE (e^{\lambda (Z - \eE Z)}) $. \text{\ding{173}} is due to the Jensen inequality, and \text{\ding{174}} is due to the assumption \ref{thm:assump1}, the last inequality \text{\ding{175}} is due to Theorem 1 \cite{bousquet2002bennett,bousquet2003concentration}, i.e., $G(\lambda) \leq \psi(-\lambda)v $). We can observe that $ v_{kj} \psi (-{\frac{\lambda \omega_{kj}}{p_k q_{kj}} }) \leq \frac{v}{W} \psi (-\lambda U)$ (proof in Lemma~\ref{pro: inequality technique}),

then we have
\begin{align}
    \pP(Z-\eE Z \geq t) \leq e^{\frac{v}{W} \psi (-\lambda U) - \lambda t},
\end{align}
we solve the minimum optimization problem with respect to $\lambda$, for $\lambda = \frac{ln(1+\frac{tW}{vU})}{U}$, then
\begin{align}
    \pP (Z- \eE Z \geq t) \leq e^{-\frac{v}{W} \varphi (\frac{tW}{vU})}.
\end{align}


Then
\begin{align*}
    U_k & =  \sum _{j \in [J_k]} w_{kj} \max (1, \triangle) \\
    & \leqone \sum _{j \in [J_k]} w_{kj} (1 + \frac{v_{kj} W}{4 v}) 
     = w_k + \frac{v_k W}{4v}, \\
    U & = \sum _{k \in [K]} U_k 
     = \sum _{k \in [K]} (w_k + \frac{v_k W}{4v}) 
     = W + \frac{W}{4} = \frac{5}{4}W,
\end{align*}
\text{\ding{172}} is according to the fact that $\forall x \in \sR,x \leq 1 + \frac{x^2}{4}$, i.e., $\max (1, \triangle) \leq 1 + \frac{v_{kj} W}{4v}$. Thus
\begin{align*}
    \pP (Z \geq \eE[Z] & + t) \leq \exp \left( -\frac{v}{W} \varphi \left( \frac{tW}{Uv}\right) \right) \notag \\
                    \leq \exp & \left( - \frac{v}{ \sum_{k \in [K]} \chi_{f} (G_k)} \varphi \left( \frac{4t}{5v}\right) \right).
\end{align*}

Since $x \geq 0, \varphi (x) \geq \frac{x^2}{2+\frac{2x}{3} } $, inequality (\eqref{eq:thm_bennet_2}) is deduced.   
\end{proof}
\begin{lemma}
    \label{pro: inequality technique} If we define $p_k, q_{kj}$ as Theorem \ref{thm:bennett_inequality}, then we can get $v_{kj} \psi(-\frac{\lambda \omega_{kj}}{p_k q_{kj}} ) \leq \frac{v}{W} \psi(-\lambda U )$.    
\end{lemma}
\begin{proof}
   \begin{itemize}
    \item $\triangle \leq  1$, then 
 \begin{align*}
     q_{kj} = \frac{\omega_{kj}}{U_k}, v \geq v_{kj} W,
 \end{align*}
 then 
 \begin{align*}
     v_{kj} \psi (-\frac{\lambda \omega_{kj}}{p_k q_{kj}}) = v_{kj} \psi (- \lambda U) \leq \frac{v}{W} \psi (- \lambda U);
 \end{align*}
 \item $\triangle > 1$,then
 \begin{align*}
     q_{kj} = \frac{\omega_{kj} v_{kj}^{\frac{1}{2}} W^{\frac{1}{2}} v^{- \frac{1}{2}} }{U_k}, v < v_{kj} W,
 \end{align*}
then
\begin{align*}
    v_{kj} \psi (-\frac{\lambda \omega_{kj}}{p_k q_{kj}}) = v_{kj} \psi (- \lambda U \frac{v^{\frac{1}{2}}}{v_{kj}^{\frac{1}{2}} W^{\frac{1}{2}} }) \leq \frac{v}{W} \psi(-\lambda U).
\end{align*}
 
\end{itemize} 
\end{proof}

\subsubsection{Proof of Theorem \ref{thm:bennett_inequality_refined}}
\label{pro:theorem4_proof}
Here we give the proof of \textbf{Theorem \ref{thm:bennett_inequality_refined} (A new refined Bennett's inequality for a special case of multi-graph dependent variables)}.
\begin{proof}

Because of $\omega_{kj}=1$, we can get $\omega_k = |J_k| \in [1,m_k]$, every $\vx_{kj}$  in  only one independent set. Then we can define $p_k,q_{kj}$ and some constants as the following:
        \begin{align*}
            & \sigma^2 = \sum_{k \in [K]} \sum_{j \in [J_k]} \omega_{kj} \sigma_{kj}^2, \ v = (1 + b) \eE [Z] + \sigma^2, p_k = \frac{U_k}{U}, U = \sum _{k \in [K]} U_k, \\
            & c = \sum_{k \in [K]} \chi_{f} (G_k), \ \chi_{f} (G_k) \defeq \sum_{j \in [J_k]} \omega_{kj}, q_{kj} = \frac{1}{U_k}, U_k = \sum _{j \in [J_k]} = \omega_k .
        \end{align*}
Review the proof of the Theorem \ref{thm:bennett_inequality}, we can get:
\begin{align} \label{equ: bunnet_refined}
    \pP (Z - \eE Z \geq t) \leq e^{(\log \sum _{k \in [K]} p_k \sum _{j \in [J_k]} q_{kj} (v_{kj} \psi (-\frac{\lambda}{p_k q_{kj}})) - \lambda t)},
\end{align}
then
\begin{align*}
    v_{kj} \psi (-\frac{\lambda}{p_k q_{kj}}) = v_{kj} \psi (- \lambda U) \leqone v \psi (- \lambda U),
\end{align*}
the inequality \text{\ding{172}} is due to every $v_{kj} \geq 0$, $v = \sum _{k \in [K]} \sum _{j \in [J_k]} v_{kj} \geq v_{kj} $. Then the inequality \eqref{equ: bunnet_refined} can be written as the following:

\begin{align*}
    \pP (Z - \eE Z \geq t) \leq e^{v \psi(- \lambda U) - \lambda t}.
\end{align*}
We solve the minimum optimization problem with respect to $\lambda$, for $\lambda = \frac{\ln (1 + \frac{t}{v U})}{U}$, then
\begin{align*}
    \pP (Z - \eE Z \geq t) \leq e^{-v \varphi (\frac{t}{v U})}.
\end{align*}
Since $U = \sum _{k \in [K]} U_k = \sum _{k \in [K]} \omega _k = W  $, and we noticed $W = \sum _{k \in [K]} \chi_f (G_k)$. Finally, we can get the one part in Theorem \ref{thm:bennett_inequality_refined}, and the second part in Theorem \ref{thm:bennett_inequality_refined} is due to the fact $x \geq 0$, $\varphi (x) \geq \frac{x^2}{2 + \frac{2x}{3}}$.
\end{proof}

\subsubsection{Proof of Theorem \ref{thm:the core 2.1}}
\label{pro:core2.1_proof}
Here we give the proof of \textbf{Theorem \ref{thm:the core 2.1} (A risk bound of multi-graph dependent variables with small variance)}.

\begin{proof}
We can define some variables, $V^+ = \sup_{f \in \mathcal{F}} (P f - P_m f)$, $\mathcal{F}_r = \{f, f \in \mathcal{F}, \mathrm{var}(f_k) \leq r\}$, 
and $f = \sum _{k \in [K]} \sum _{j \in [J_k]} \omega_{kj} \sum _{i \in I_{kj}} f_k(\vx_i) $.
then
\begin{align*}
    V^+ = \sup _{f \in \mathcal{F}_r} (P f - P _m f) & = \sup _{f \in \mathcal{F}_r} \eE _{\vx'} [ (\frac{1}{K} \sum _{k \in [K]} \sum _{j \in [J_k]} \frac{\omega_{kj}}{m_k} \sum _{i \in I_{kj}}  f_k(\vx'_i) - \frac{1}{K} \sum _{k \in [K]} \sum _{j \in J_{k}} \frac{\omega_{kj}}{m_k} \sum _{i \in I_{kj}} f_k(\vx_i))] \\
    & \leq \eE _{\vx'} [ \sup _{f \in \mathcal{F}_r}   (\frac{1}{K} \sum _{k \in [K]} \sum _{j \in [J_k]} \frac{\omega_{kj}}{m_k} \sum _{i \in I_{kj}}  f_k(\vx'_i) - \frac{1}{K} \sum _{k \in [K]} \sum _{j \in J_{k}} \frac{\omega_{kj}}{m_k} \sum _{i \in I_{kj}} f_k(\vx_i)) ] \\
    & = \eE _{\vx'} [ \sup _{f \in \mathcal{F}_r}   (\frac{1}{K} \sum _{k \in [K]} \sum _{j \in [J_k]} \frac{\omega_{kj}}{m_k} \sum _{i \in I_{kj}} ( f_k(\vx'_i) - f_k(\vx_i) ) ], \\
\end{align*}
which has differences bounded by $ \frac{1}{K m_k} $ in the sense of the Z in Theorem \ref{thm:talagrand_inequality}, then with probability at least $ 1 - e^{-t}$, 
\begin{align}
    V^+ \leq \eE V^+ + \frac{1}{K} \sqrt{2cvt} + \frac{2ct}{3K},    
\end{align}
where $ c = \frac{5^2}{4^2} \sum _{k \in [K]} \frac{\chi _f(G_k)}{m_k}$ and $ v = \sum _{k \in [K]} \frac{1}{m_k} \sum _{j \in J_k} \omega_{kj} \sigma _{kj}^2 +2K\eE V^+ \leq Kr + 2K\eE V^+ $. 
Then 
\begin{align*}
    V^+ & \leq \eE V^+ + \sqrt{\frac{2c(r + 2\eE V^+) t}{K}} +\frac{2ct}{3K} \\
    & \leqone \eE V^+ +\sqrt {\frac{4c \eE V^+ t}{K}} + \sqrt {\frac{2crt}{K}} + \frac{2ct}{3K} \\
    & \leqtwo (1+\alpha) \eE V^+ + \sqrt {\frac{2crt}{K}} + (\frac{2}{3} + \frac{1}{\alpha})\frac{ct}{K} \\  
    & \leqthree 2(1 + \alpha) \mathcal{R} (\mathcal{F} , r) + \sqrt {\frac{2crt}{K}} + (\frac{2}{3} + \frac{1}{\alpha})\frac{ct}{K}, \\
\end{align*}

where \text{\ding{172}} is due to the fact $\sqrt{a + b} \leq \sqrt{a} + \sqrt{b}$, i.e., $\sqrt{\frac{2c(r + 2\eE V^+) t}{K}} \leq \sqrt {\frac{4c \eE V^+ t}{K}} + \sqrt {\frac{2crt}{K}} $, and \text{\ding{173}} is due to the fact $\forall \alpha > 0, 2 \sqrt{ab} \leq \frac{a}{\alpha} + \alpha b$,  and we can combine similar items. \text{\ding{174}} is due to  the proposition \ref{pro: equ Rademacher}. 
\end{proof}

\subsubsection{Proof of Theorem \ref{thm: theorem 3.3 sub-root}}
\label{pro:theorem3.3_proof}
Here we give the proof of \textbf{Theorem \ref{thm: theorem 3.3 sub-root} (An improved bound of multi-graph dependent variables with sub-root function)}.
\begin{proof} 
We can define $\mathcal{G}_r = \{ g = (g_1,g_2,... ,g_K), g_k = \frac{r}{w(f_k)}f_k, f \in \mathcal{F} \} $,  $\mathcal{F}(x,y): = \{ f \in \mathcal{F}, T(f_k) \in [x,y] \} $, where $w(f_k) = \min \{r\lambda ^a, a \in \mathbb{N}, r\lambda ^a \geq T(f_k), f \in \mathcal{F} \}$, $V_r^+ = \sup _{g \in \mathcal{G}_r} (P g - P _m g)$. We can notice $\frac{r}{w(f_k)} \in [0,1]$, then for each $g \in \mathcal{G}_r,  \| g_k\|_ \infty \leq 1$, and we found $\mathrm{var}(g_k) \leq r$. Because

\begin{itemize}
    \item $T(f_k) \leq r$,
then
    \begin{align*}
        a = 0, w(f) = 1,
        \forall g \in \mathcal{G}_r, g = f ,
        \mathrm{var}(g_k) = \mathrm{var}(f_k) \leq r;
    \end{align*}

    \item $T(f_k) >r$,
then
    \begin{align*}
        g = \frac{f}{\lambda^a}, T(f_k) \in (r \lambda^{a-1}, r\lambda^a], 
        \mathrm{var}(g_k) = \frac{\mathrm{var}(f_k)}{\lambda^{2a}} \leq r.
    \end{align*}
\end{itemize}

Then we can apply Theorem \ref{thm:the core 2.1} for $\mathcal{G}_r$, for all $x > 0$, with probability $1 - e^{-t}$,
\begin{align*} 
    V_r^+ \leq 2(1+\alpha) \mathcal{R}(\mathcal{G}_r) +\sqrt{\frac{2crt}{K}} + (\frac{2}{3} +\frac{1}{\alpha}) \frac{ct}{K}.   
\end{align*}


Then

\begin{align*}
    \mathcal{R}(\mathcal{G}_r) = \frac{1}{K} \left[ \eE_ \vx  \eE _\zeta \underbrace{  \sup _{g \in \mathcal{G}_r,\mathrm{var}(g_k) \leq r} \sum_{k \in [K]} \frac{1}{m_k}  \sum_{j \in J_k} \omega_{kj}  \sum _{i \in I_{kj}} \zeta_l g_k(x_i) }_\spadesuit \right].   
\end{align*}

Let $T(f_k) \leq Bb$, define $a_0$ to be the smallest integer that $r\lambda^{a_0+1} \geq Bb $ and partition $\mathcal{F}(0,Bb) $, i.e., $\mathcal{F}(0,r) + \mathcal{F}(r\lambda^0,r \lambda^1) +\mathcal{F}(r\lambda^1,r \lambda^2),...(r\lambda^{a_0}, r \lambda^{a_0+1}) $, then
    

\begin{align*}
    \spadesuit & \leq \sup_{g \in \mathcal{G}_r, T(f_k) \in [0,r]}  \sum_{k \in [K]} \frac{1}{m_k} \sum_{j \in J_k} \omega_{kj} \sum _{i \in I_{kj}} \zeta_i g_k(x_i) +\sup _{g \in \mathcal{G}_r, T(f) \in [r,Bb]} \sum_{k \in [K]} \frac{1}{m_k} \sum_{j \in J_k} \omega_{kj} \sum _{i \in I_{kj}} \zeta_i g_k(x_i) \\
    & = \sup _{f \in \mathcal{F}(0,r)} \sum_{k \in [K]} \frac{1}{m_k} \sum_{j \in J_k} \omega_{kj} \sum _{i \in I_{kj}} \zeta_i f_k(x_i) + \sup _{f \in \mathcal{F}(r,Bb)} \sum_{k \in [K]} \frac{1}{m_k} \sum_{j \in J_k} \omega_{kj} \sum _{i \in I_{kj}} \zeta_i \frac{r}{w(f_k)} f_k(x_i) \\
    & \leq \sup _{f \in \mathcal{F}(0,r)} \sum_{k \in [K]} \frac{1}{m_k} \sum_{j \in J_k} \omega_{kj} \sum _{i \in I_{kj}} \zeta_i f_k(x_i) + \sum _{j=0} ^{a_0} \sup _{f \in \mathcal{F}(r\lambda ^j, r \lambda ^{j+1})} \sum_{k \in [K]} \frac{1}{m_k} \sum_{j \in J_k} \omega_{kj} \sum _{i \in I_{kj}} \zeta_i \frac{r}{r\lambda^j} f_k(x_i) \\
    & = \sup _{f \in \mathcal{F}(0,r)} \sum_{k \in [K]} \frac{1}{m_k} \sum_{j \in J_k} \omega_{kj} \sum _{i \in I_{kj}} \zeta_i f_k(x_i) + \sum _{j=0} ^{a_0} \frac{1}{\lambda^j} \sup _{f \in \mathcal{F}(r\lambda ^j, r \lambda ^{j+1})} \sum_{k \in [K]} \frac{1}{m_k} \sum_{j \in J_k} \omega_{kj} \sum _{i \in I_{kj}} \zeta_i f_k(x_i).
\end{align*}

Since the property in Definition~\ref{lemma: sub-root def}, i.e., $\forall \gamma \geq 1, \Phi(\gamma r) \leq \sqrt{\gamma} \Phi(r)$,  then we can get: 
\begin{align*}
    \mathcal{R}(\mathcal{G}_r) & \leq \mathcal{R} \mathcal{F} (0,r) + \sum _{j=0} ^{a_0} \frac{1}{\lambda^j} \mathcal{R} \mathcal{F}(r \lambda^j, r \lambda^{j+1} ) 
     \leq \frac{\Phi(r)}{B} + \frac{1}{B} \sum _{j=1} ^{a_0} \frac{1}{\lambda^j} \Phi (r \lambda^{j+1}) \\
    & \leq \frac{\Phi(r)}{B} +  \frac{1}{B} \sum _{j=1} ^{a_0} \frac{1}{\lambda^j} \lambda^{\frac{j+1}{2}} \Phi (r) 
     = \frac{\Phi(r)}{B} \left[ 1 + \sqrt{\lambda} \sum _{j=0}^{a_0} \frac{1}{\sqrt{\lambda^j}}   \right],
\end{align*}
then taking $\lambda = 4$, we can get: 
\begin{align*}
    \mathcal{R}(\mathcal{G}_r) \leq \frac{5\Phi(r)}{B}.
\end{align*}
Since the property in Definition~\ref{lemma: sub-root def} and Lemma \ref{lemma:sub-root pro}, i.e., $\Phi(r^*) = r^*$, and $\frac{\Phi(r)}{\sqrt{r}}$ is nonincreasing, thus
\begin{align*}
    V_r^+ & \leq 2(1+\alpha) \frac{5\Phi(r)}{B} +\sqrt{\frac{2crt}{K}} + (\frac{2}{3} +\frac{1}{\alpha}) \frac{ct}{K} \\
    & \leq \frac{10(1+\alpha)}{B} \sqrt{r r^*} + \sqrt{\frac{2crt}{K}} + (\frac{2}{3} +\frac{1}{\alpha}) \frac{ct}{K}.
\end{align*}
Then let $A = \frac{10(1+\alpha)}{B} \sqrt{r^*} + \sqrt{\frac{2ct}{K}} $, $C = (\frac{2}{3} +\frac{1}{\alpha}) \frac{ct}{K} $, thus
\begin{align*}
    V_r^+ \leq A \sqrt{r} + C.
\end{align*}

Then we can apply lemma 3.8 in \cite{Bartlett_2005}, i.e., if $V_r^+ \leq \frac{r}{\lambda B M }$, then $P f \leq \frac{M}{M-1} P_m f + \frac{r}{\lambda BM}$, where $\lambda = 4$, and use a technique, i.e., $\forall \beta >0 , \sqrt{ab} \leq \frac{1}{2}(\beta a + \frac{b}{\beta})$, then
\begin{align*}
    P f & \leq \frac{M}{M-1} P _m f + \lambda BMA^2 +2C \\
    & = \frac{M}{M-1} P _m f + \lambda B M (\frac{100(1+\alpha)^2 r^*}{B^2} + \frac{20(1+\alpha)}{B} \sqrt{\frac{2cr^*t}{K}} + \frac{2ct}{K}) + (\frac{2}{3}+\frac{1}{\alpha}) \frac{2ct}{K}, \\
     & \leq \frac{M}{M-1} P _m f + \lambda B M \left[\frac{100(1+\alpha)^2 r^*}{B^2} + \frac{20(1+\alpha)}{B} \left(\frac{1}{2} ( \frac{5r^*}{B} +  \frac{2cBt}{5K}) \right) + \frac{2ct}{K}  \right] + (\frac{2}{3}+\frac{1}{\alpha}) \frac{2ct}{K}, \\
\end{align*}
let $\alpha = \frac{1}{10}$, then
\begin{align*}
    P f & = \frac{M}{M-1} P _m f + \lambda B M (\frac{121 r^*}{B^2} + \frac{22}{B} (\frac{cBt}{5K} + \frac{5r^*}{2B}) + \frac{2ct}{K}) + \frac{32}{3} \frac{2ct}{K} \\
    & = \frac{M}{M-1} P _m f + \frac{(121+55)\lambda M}{B} r^* + \left( \frac{32 \lambda B M}{5} + \frac{64}{3} \right) \frac{ct}{K} \\
    & \leq \frac{M}{M-1} P _m f + \frac{704 M}{B} r^* + (26 BM +22 ) \frac{ct}{K}, \\
\end{align*}
where $c = \frac{5^2}{4^2} \sum _{k \in [K]} \frac{\chi_f(G_k)}{m_k}$. Let $c_1 = 704$, $c_2 = 26$, then we can get inequality \eqref{eq: thm sub-root 3.3_1}. In the same way, we can define $V_r^- = \sup _{g \in \mathcal{G}_r} (P _m g - P g)$, and then get 

\begin{align} \label{eq: thm sub-root 3.3_2}
     P_m f \leq \frac{M}{M-1} P  f + \frac{c_1 M}{B} r^* + (c_2 BM +22 ) \frac{ct}{K},
\end{align}
\end{proof}

\subsubsection{Proof of Proposition \ref{thm:kernel upper bound}}
\label{pro:proposition1_proof}
Here we give the proof of \textbf{Proposition \ref{thm:kernel upper bound} (The upper bound of LFRC in kernel hypothesis)}. 
\begin{proof} 
We can  observe that   $    \mathcal{R}\{ f \in \mathcal{F}, \eE f_k^2 \leq r \} = \mathcal{R}\{ f \in \mathcal{F}, \eE(\|f_k\|_2) \leq \sqrt{r} \} \leq \mathcal{R}\{f \in \mathcal{F}, \eE (\|f_k\|_1) \leq \sqrt{r}\}$.
Donate $\mathcal{R}(\mathcal{F}) \leq \mathcal{R} (\mathcal{F}_{2,1})$, $\mathcal{F}_{2,1} = \{ f = (f_1,f_2,...,f_k), k \in [K], \eE (\|f_k\|)_1 \leq \sqrt{r} \}$, and $\mathcal{F}_{k_{2,1}}
 \{ f_k : \vx \rightarrow \Theta^T \phi (\vx), \|\theta_k\|_2 \leq Ma, \eE(\|f_k\|_1) \leq \sqrt{r} \}$. Then we fix $d = (d_1,d_2,...,d_K)$, 
\begin{align*}
    & \frac{1}{K} \sum _{k \in [K]} \frac{1}{m_k} \sum _{j \in [J_k]} \omega_{kj} \sum _{i \in I_{kj}} \zeta_i < \theta_k, \phi (\vx_i)> \\
    = & \sum _{k \in [K]} \frac{\chi_f(G_k)}{K m_k} \sum _{j \in [J_k]} \frac{\omega_{kj}}{\chi_f(G_k)} \sum _{i \in [I_{kj}]} \zeta_i <\theta_k, \phi(\vx_i)> \\
     = &  \sum _{k \in [K]}  <\theta_k, \frac{\chi _f(G_k)}{K m_k} \sum _{j \in [J_k]} \frac{\omega_{kj}}{\chi _f(G_k)} \sum _{i \in I_{kj}} \zeta_i \phi (\vx_i)> \\  
     \leq & \sum _{k \in [K]}  <\theta_k, \frac{\chi _f(G_k)}{K m_k} \zeta_k \phi(\vx_k) > \\
     =  & \sum _{k \in [K]} [ <\sum _{l = 1}^{d_k} \sqrt{\lambda _{kl}}<\theta_k, \varphi _{kl}>\varphi _{kl} , \sum _{l = 1}^{d_k} \frac{1}{\sqrt{ \lambda _{kl}}} <\frac{\chi_f(G_k)}{K m_k} \zeta_k \phi (\vx_k) , \varphi _{kl} > \varphi _{kl}> \\ + & <\theta_k, \sum _{l > d_k} <\frac{\chi_f(G_k)}{K m_k} \zeta_k \phi (\vx_k), \varphi _{kl} >, \varphi _{kl}>] \\
\end{align*}
Let $\diamondsuit = \frac{1}{K} \sum _{k \in [K]} \frac{1}{m_k} \sum _{j \in [J_k]} \omega_{kj} \sum _{i \in I_{kj}} \zeta_i <\theta_k, \phi (\vx_i)> $,
\begin{align*}
    \eE \sup_{f \in \mathcal{F}} (\diamondsuit) & \leq \sup_{ \|\theta_k\|_2 \leq Ma } \sum _{k \in [K]} \sqrt{(\sum _{l=1}^{d_k} \lambda_{kl} <\theta_k, \varphi _{kl}>^2) (\frac{\chi_f(G_k)}{K m_k} \sum _{l=1}^{d_k} \frac{1}{\lambda _{kl}} \eE [< \zeta_k \phi (\vx_k), \varphi _{kl}>^2])} \\
    & + \|\theta_k\|_2 \sqrt{\frac{\chi_f(G_k)}{K m_k} \sum_{l > d_k} \eE[ < \zeta_k \phi (\vx_k), \varphi _{kl}>^2 ]}, \\
\end{align*}
since $\sum _{l=1}^{d_k} \lambda_{kl} <\theta_k, \varphi _l>^2 \leq r$,   $\eE [< \zeta_k \phi (\vx_k), \varphi _{kl}>^2] = \lambda_{kl}$, then
\begin{align*}
    \eE \sup_{f \in \mathcal{F}} (\diamondsuit) \leq \sum _{k \in [K]} \sqrt{\frac{r d_k \chi_f(G_k)}{K m_k}} + \widetilde{M} \sqrt{\frac{\chi_f(G_k)}{K m_k} \sum _{l > d_k} \lambda _{kl} }.\\
\end{align*}

So $\mathcal{R}(\mathcal{F}) \leq \mathcal{F}_{2,1} \leq \sum _{k \in [K]} \min _{d_k \geq 0} \sqrt{\frac{r d_k \chi_f(G_k)}{K m_k}} + \widetilde{M} \sqrt{\frac{\chi_f(G_k)}{K m_k} \sum _{l > d_k} \lambda _{kl}}$.
\end{proof}

\begin{proposition} [The upper bound of LFRC in kernel hypothesis with special case] \label{thm:kernel compute2}
Assume for each $k \in [K]$, $\|\theta_k\|_2 \leq \widetilde{M}$. For all $r > 0$,  
\begin{align}
    \mathcal{R}\{ f \in \mathcal{F}, \eE f_k^2 \leq r \} \leq \left(\sum _{k \in [K]} \frac{2 \chi_f(G_k)}{m_k} \sum_{l=1}^{\infty} \min \{r,\widetilde{M}^2 \lambda_l \} \right)^\frac{1}{2},
\end{align}
where eigenvalues $(\lambda _l) _{l=1} ^{\infty}$ are in a nonincreasing order, and satisfy $\kappa (\vx,\vx') = \sum _{l =1} ^{\infty} \lambda_l \varphi_l (\vx)^T \varphi _l (\vx') $. In this theorem, we consider the data for all tasks to share a kernel matrix, which is used more commonly in practical applications. 
    
\end{proposition}
\begin{proof} 
We can use a similar method (eigenvalue decomposition) to the Proposition \ref{thm:kernel upper bound}, then
    \begin{align*}
       & \frac{1}{K} \sum _{k \in [K]} \frac{1}{m_k} \sum _{j \in J_k} \omega_{kj} \sum _{i \in I_{kj}} \zeta_i < \theta_k, \phi (\vx_i)> \\
  \leq & \sum _{k \in [K]} [ <   \sum _{l = 1}^{d} \sqrt{\lambda _{l}}<\theta_k,  \varphi _{l}>\varphi _{l} ,  \sum _{l = 1}^{d} \frac{1}{\sqrt{ \lambda _{l}}} <\frac{\chi_f(G_k)}{K m_k}  \zeta_k \phi (\vx_k) , \varphi _{l} > \varphi _{l}  > \\
  + & <  \theta_k,  \sum _{l > d} <\frac{1}{K m_k}  \zeta_k \phi (\vx_k), \varphi _{l} >, \varphi _{l} > ].
    \end{align*} 
Let $\Box = \frac{1}{K} \sum _{k \in [K]} \frac{1}{m_k} \sum _{j \in J_k} \omega_{kj} \sum _{i \in I_{kj}} \zeta_i < \theta_k, \phi (\vx_i)>$, then
    \begin{align*}
    \eE \sup_{f \in \mathcal{F}} (\Box) & \leq \sup_{ \|\theta_k\|_2 \leq Ma } \sum _{k \in [K]} \sqrt{(\sum _{l=1}^{d} \lambda_l <\theta_k, \varphi _l>^2) (\frac{\chi_f(G_k)}{K m_k} \sum _{l=1}^{d} \frac{1}{\lambda _l} \eE [< \zeta_k \phi (\vx_k), \varphi _l>^2])} \\
    & + \|\theta_k\|_2 \sqrt{\frac{\chi_f(G_k)}{K m_k} \sum_{l > d} \eE[ < \zeta_k \phi (\vx_k), \varphi _l>^2 ]} \\
    & \leq \sup_{\|\theta_k\|_2 \leq \widetilde{M}} \sum_{k \in [K]} \sqrt{\sum_{l=1}^d \lambda_l<\theta_k, \varphi_l>^2} \cdot \sum_{k \in [K]} \sqrt{ (\frac{\chi_f(G_k)}{K m_k} \sum _{l=1}^{d} \frac{1}{\lambda _l} \eE [< \zeta_k \phi (\vx_k), \varphi _l>^2])} \\
    & + \widetilde{M} \sum_{k \in [K]} \sqrt{\frac{\chi_f(G_k)}{K m_k} \sum_{l > d} \eE[ < \zeta_k \phi (\vx_k), \varphi _l>^2 ]} \\
    & \leq \sup_{\|\theta_k\|_2 \leq \widetilde{M}} \sum_{k \in [K]} \sqrt{\sum_{l=1}^d \lambda_l<\theta_k, \varphi_l>^2} \cdot \sqrt{K} \sqrt{ \sum_{k \in [K]} (\frac{\chi_f(G_k)}{K m_k} \sum _{l=1}^{d} \frac{1}{\lambda _l} \eE [< \zeta_k \phi (\vx_k), \varphi _l>^2])} \\
    & + \widetilde{M} \sqrt{K} \sqrt{ \sum_{k \in [K]} \frac{\chi_f(G_k)}{K m_k} \sum_{l > d} \eE[ < \zeta_k \phi (\vx_k), \varphi _l>^2 ]}\\
    & \leq \sup_{\|\theta_k\|_2 \leq \widetilde{M}}  \sum_{k \in [K]} \sqrt{\sum_{l=1}^d \lambda_l<\theta_k, \varphi_l>^2} \cdot \sqrt{ \left( \sum_{k \in [K]} \frac{\chi_f(G_k)}{m_k} \right) \left( \sum_{k \in [K]} \sum _{l=1}^{d} \frac{1}{\lambda _l} \eE [< \zeta_k \phi (\vx_k), \varphi _l>^2] \right) } \\
    & + \widetilde{M}  \sqrt{ \left( \sum_{k \in [K]} \frac{\chi_f(G_k)}{m_k} \right) \left( \sum_{k \in [K]} \sum_{l > d} \eE[ < \zeta_k \phi (\vx_k), \varphi _l>^2 ] \right) }\\
    & \leq \sup_{\|\theta_k\|_2 \leq \widetilde{M}}  \sum_{k \in [K]} \sqrt{\sum_{l=1}^d \lambda_l<\theta_k, \varphi_l>^2} \cdot \sqrt{ \left( \sum_{k \in [K]} \frac{\chi_f(G_k)}{m_k} \right) \left( \sum _{l=1}^{d} \frac{1}{\lambda _l} \eE [< \phi(\vx), \varphi _l>^2] \right) } \\
    & + \widetilde{M}  \sqrt{ \left( \sum_{k \in [K]} \frac{\chi_f(G_k)}{m_k} \right) \left( \sum_{l > d} \eE[ <  \phi (\vx), \varphi _l>^2 ] \right) }.    
\end{align*}

Since $\sum_{k \in [K]} \sqrt{\sum_{l=1}^d \lambda_l<\theta_k, \varphi_l>^2} \leq \sqrt{r}$, $\eE [<\phi(\vx), \varphi_l>^2] = \lambda_l$, then
\begin{align*}
    \eE \sup_{f \in \mathcal{F}} (\diamondsuit) \leq \sqrt{\sum _{k \in [K]} \frac{\chi_f(G_k)}{m_k} r d } +\widetilde{M} \sqrt{\sum_{k \in [K]} \frac{\chi_f(G_k)}{m_k} \sum_{l > d} \lambda_l },
\end{align*}
so $\mathcal{R}(\mathcal{F}) \leq \min_{d \geq 0} \sqrt{\sum _{k \in [K]} \frac{\chi_f(G_k)}{m_k} r d } +\widetilde{M} \sqrt{\sum_{k \in [K]} \frac{\chi_f(G_k)}{m_k} \sum_{l > d} \lambda_l } $.   
\end{proof}

\subsubsection{Proof of Corollary \ref{thm: loss bound computing}}
\label{pro:corollary2.1_proof}
Here we give the proof of \textbf{Corollary \ref{thm: loss bound computing} (A risk bound of LFRC in kernel hypothesis)}.
\begin{proof}  We can notice that 
\begin{align*}
    \mathcal{R}\{ h \in \mathcal{H}, \mu^2 \eE (h_k - h_k^*)^2 \leq r \} & = \mathcal{R}\{ h \in \mathcal{H}, \eE (h_k - h_k^*)^2 \leq \frac{r}{\mu ^2} \} 
     = \mathcal{R}\{ h-h^*, h \in \mathcal{H}, \eE (h_k - h_k^*)^2 \leq \frac{r}{\mu ^2} \} \\
    & \leq \mathcal{R}\{ h - g, h,g \in \mathcal{H}, \eE (h_k - g_k^*)^2 \leq \frac{r}{\mu ^2} \} 
     = 2 \mathcal{R}\{ h, h \in \mathcal{H}, \eE h_k ^2 \leq \frac{r}{4 \mu ^2} \},
\end{align*}
We define a function $\Psi(r) = \frac{\widetilde{c}_1}{2} \mathcal{R}\{ h \in \mathcal{H}, \mu^2\eE (h_k - h_k^*)^2 \leq r \}+\frac{11\mu^2 t}{m}$, $\widetilde{c}_1 = 2B\mu$. $\Psi(r)$ is a sub-root function (owing to Lemma~3.4 in \cite{Bartlett_2005}).
Then use the Proposition~\ref{thm:kernel upper bound}, we can get
\begin{align*}
    \Psi(r) \leq \sum_{k \in [K]} \left( \frac{2 \chi_f(G_k)}{K m_k} \sum_{l=1}^{\infty} \min\{\frac{r}{4 \mu^2},\widetilde{M}^2 \lambda_{kl}\} \right)^{\frac{1}{2}} + \frac{11 \mu^2 t}{m}.
\end{align*}
Similar to \cite{Bartlett_2005} (i.e., adding a constant $c$ to a sub-root function can increase its fixed point by at most $2c$), it suffices to show that 
\begin{align*}
    r \leq 2B\mu \sum_{k \in [K]} \left( \frac{2 \chi_f(G_k)}{K m_k} \sum_{l=1}^{\infty} \min\{\frac{r}{4\mu^2},\widetilde{M}^2 \lambda_{kl}\} \right)^{\frac{1}{2}}.
\end{align*}
Then 
\begin{align*}
    r \leq C \left[ \sum_{k \in [K]} \min _{0 \leq d_k \leq m_k} \left( \frac{1}{2 \mu} \sqrt{\frac{r d_k \chi_f(G_k)}{K m_k}} +\widetilde{M} \sqrt{\frac{\chi_f(G_k)}{K m_k} \sum _{l > d_k} \lambda _{kl}} \right) \right].
\end{align*}
Since $r \geq \Psi(r)$ and
the property in Lemma \ref{lemma:sub-root pro}, i.e., the fixed point $r^*$ satisfied $r^* = \Psi(r^*)$. 
Then 
\begin{align*}
    r ^* \leq C \left[ \sum_{k \in [K]} \min _{0 \leq d_k \leq m_k} \left( \frac{1}{2 \mu} \sqrt{\frac{r d_k \chi_f(G_k)}{K m_k}} +\widetilde{M} \sqrt{\frac{\chi_f(G_k)}{K m_k} \sum _{l > d_k} \lambda _{kl}} \right) \right],
\end{align*}
where $C$ is a constant about $B, \mu$. 

\end{proof}

\subsubsection{Proof of Proposition \ref{thm:linear upper bound}}
\label{pro:proposition3_proof}
Here we give the proof of \textbf{Proposition \ref{thm:linear upper bound} (The upper bound of LFRC in linear hypothesis)}.
\begin{proof} 
Similarly, we can get $\mathcal{R}\{ f \in \mathcal{F}, \eE f_k^2 \leq r \}  = \mathcal{R} \{ f \in \mathcal{F}, \eE [\vx^T \theta_k \theta_k^T \vx] \leq r \} 
    = \mathcal{R} \{ f \in \mathcal{F}, \eE [\| \theta_k \theta_k^T \|] \leq \frac{\sqrt{r}}{\bar{M}} \}$.  
Then we donate the above as $\mathcal{R}(\mathcal{F}) = \mathcal{R}(\mathcal{F}_{2,2})$, and consider the SVD composition of $\Theta$, i.e., $\Theta = \sum_{l = 1} u_l v_l^T \widetilde{\lambda}_l $, where ${\{\widetilde{\lambda}_l\}}_{l = 1}^{\infty}$ are the singular values of $\Theta$ and are sorted in a nonincreasing order. Then
\begin{align*}
     & \frac{1}{K} \sum _{k \in [K]} \frac{1}{m_k} \sum _{j \in [J_k]} \omega_{kj} \sum _{i \in I_{kj}} \zeta _i <\theta_k, \vx_i> 
    = \sum_k \frac{\chi_f(G_k)}{K m_k} \sum_{j \in [J_k]} \frac{\omega_{kj}}{\chi_f(G_k)} \sum_{i \in [I_{kj}]} \zeta_i < \theta_k, \vx_i > \\
    = &  \sum _{k \in [K]}  <\theta_k, \frac{\chi _f(G_k)}{K m_k} \sum _{j \in [J_k]} \frac{\omega_{kj}}{\chi _f(G_k)} \sum _{i \in I_{kj}} \zeta_i \vx_i> \\  
     \leq & \sum _{k \in [K]}  <\theta_k, \frac{\chi _f(G_k)}{K m_k} \zeta_k \vx_k > \\
     =  & \sum _{k \in [K]} [ <\sum _{l = 1}^{d_k} \sqrt{\lambda _{kl}}<\theta_k, \varphi _{kl}>\varphi _{kl} , \sum _{l = 1}^{d_k} \frac{1}{\sqrt{ \lambda _{kl}}} <\frac{\chi_f(G_k)}{K m_k} \zeta_k \phi (\vx_k) , \varphi _{kl} > \varphi _{kl}> \\ + & <\theta_k, \sum _{l > d_k} <\frac{\chi_f(G_k)}{K m_k} \zeta_k \vx_k, \varphi _{kl} >, \varphi _{kl}>] \\ 
\end{align*}
Let $\diamondsuit = \frac{1}{K} \sum _{k \in [K]} \frac{1}{m_k}$ $ \sum _{j \in [J_k]} \omega_{kj} \sum _{i \in I_{kj}} \zeta _i <\theta_k, \vx_i>$, then similar to the proof of Proposition~\ref{thm:kernel compute2},
$\| \sum _{l=1}^{d} u_{l} v_{l}^T \widetilde{\lambda} _{l}^2 \| \leq \frac{\sqrt{r}}{\widetilde{M}}$, we can get
\begin{align*}
    \mathcal{R}\{ f \in \mathcal{F}, \eE f_k^2 \leq r \} \leq  \sum_k \left( \frac{2 \chi_f(G_k)}{K m_k} \sum _{l = 1} ^{\infty} \min \{ \frac{r}{\bar{M}^2} , \widetilde{M}^2 \widetilde{\lambda} ^2_{l} \} \right)^{\frac{1}{2}}.
\end{align*}
\end{proof}

\subsubsection{Proof of Corollary \ref{thm: loss bound computing2}}
\label{pro:corollary6_proof}
Here we give the proof of \textbf{Corollary \ref{thm: loss bound computing2} (An excess risk bound of loss space in linear hypothesis)}.
\begin{proof} 
Similarly, we can prove the Corollary \ref{thm: loss bound computing2} likely the Corollary \ref{thm: loss bound computing}.
\begin{align*}
    \mathcal{R}\{ h \in \mathcal{H}, \mu^2 \eE (h_k - h_k^*)^2 \leq r \} & = \mathcal{R}\{ h \in \mathcal{H}, \eE (h_k - h_k^*)^2 \leq \frac{r}{\mu ^2} \} 
     = \mathcal{R}\{ h-h^*, h \in \mathcal{H}, \eE (h_k - h_k^*)^2 \leq \frac{r}{\mu ^2} \} \\
    & \leq \mathcal{R}\{ h - g, h,g \in \mathcal{H}, \eE (h_k - g_k)^2 \leq \frac{r}{\mu ^2} \} 
     = 2 \mathcal{R}\{ h, h \in \mathcal{H}, \eE h_k ^2 \leq \frac{r}{4 \mu ^2} \}.
\end{align*}
Owing to the property in Lemma \ref{lemma:sub-root pro}, i.e.,  the fixed point $r^*$ satisfies $r^* = \Psi(r^*)$, Then we can use the Corollary \ref{thm : Lipschitz bound loss space} and Proposition \ref{thm:linear upper bound} to get the results. According to the Proposition \ref{thm:linear upper bound}, 
\begin{align*}
    r ^* \leq C \left[ \sum_{k \in [K]} \min _{d_k \geq 0} \left( \frac{1}{2 \mu} \cdot \frac{1}{\bar{M}} \sqrt{\frac{r d_k \chi_f(G_k)}{K m_k}} +\widetilde{M} \sqrt{\frac{\chi_f(G_k)}{K m_k} \sum _{l > d_k} \widetilde{ \lambda} _{kl}^2} \right) \right],
\end{align*}
where $C$ is a constant about $B, \mu$. If we use the second decomposition of $\Theta$, then 
\begin{align*}
    r ^* \leq C \sum_k\left[\min _{d \geq 0} \left( \frac{1}{2 \mu} \cdot \frac{1}{\bar{M}} \sqrt{ \frac{\chi_f(G_k)}{K m_k} r d}+\widetilde{M} \sqrt{ \frac{\chi_f(G_k)}{K m_k} \sum _{l > d} \widetilde{ \lambda} _{l}^2} \right) \right].
\end{align*}
Furthermore, $r^*$ satisfies
\begin{align}
    r^* \leq \sum_k \min _{d \geq 0} \left( \frac{d}{\bar{M}^2} \frac{\chi_f(G_k) }{K m_k} + \widetilde{M} \sqrt{ \frac{\chi_f(G_k)}{K m_k} \sum _{l >d} \widetilde{ \lambda} _{l}^2} \right).
\end{align}
\end{proof}

\section{Other Applications} \label{section: E}
This section provides further applications of Section \ref{section: applications}, which includes detailed theoretical results for Macro-AUC Optimization and AUUC-maximization.


    

\subsection{Macro-AUC -Optimization in MTL}
Here, we provide additional information on the theoretical results of Macro-AUC Optimization presented in the main text.
\label{pro:macro-auc-appendix}
\begin{theorem}[The base theorem of Macro-AUC, proof in Appendix \ref{section: A}]\label{thm: base auc} Assume that the loss function $L : \mathcal{X} \times \mathcal{X} \times \mathcal{H}_k \rightarrow R_+ $ is bounded by $M_c$, and $\mathcal{\tilde{H}}_ {l,r} = \{ \tilde{h}: \tilde{h} \in \mathcal{\tilde{H}}, \mathrm{var}( L_{\tilde{h}_k}) \leq r \}$, where $\widetilde{\vx}_{kl} = (\widetilde{\vx}_{kl}^+, \widetilde{\vx}_{kl}^-)$. For each $h \in \mathcal{H}_{l,r}$, $\alpha > 0 $ and $t > 0$,  with probability at least $1 - e^{-t}$, 
\begin{align*}
   & P (L_h) - P _m (L_h) \leq  2(1 + \alpha) \mathcal{R}( \mathcal{H}_{l,r}) + \\
   & \frac{5}{4} \sqrt{\frac{2rt}{\tilde{n}} \cdot \frac{1}{K} \sum _{k \in [K]} \frac{1}{\tau_k}} +  \frac{5^2}{4^2} (\frac{2}{3} + \frac{1}{\alpha}) (\frac{1}{K} \sum_{k \in [K]} \frac{1}{\tau_k}) \frac{t}{\tilde{n}},
\end{align*}
where $P(L_{\tilde{h}}), P_m(L_{\tilde{h}})$ can be obtained by Eq.\eqref{def: pmg}, i.e., $P_m(L_{\tilde{h}}) = \frac{1}{K}\sum_{k \in [K]} \sum_{j \in [J_k]}\frac{\omega_{kj}}{m_k} \sum_{i \in {I_{kj}}} L(\vx_i,y_i,\tilde{h}_k)$ and $P_{\tilde{h}} = \eE(P_m (L_{\tilde{h}}))$.
\end{theorem}

\begin{remark}
    This theorem is the basis and core of the derivation of subsequent boundaries, where we can derive more detailed generalization bounds. 
\end{remark}
The sub-root function can subsequently be introduced to derive an improved risk bound for Macro-AUC, similar to Theorem \ref{thm: theorem 3.3 sub-root}, as well as a risk bound for the bounded loss function. This development facilitates the establishment of the risk bound (see Appendix \ref{section: E}) for the loss function under Assumption \ref{thm:assump2}. Following this, generalization bounds are presented for two types of hypothesis spaces: kernel and linear. A comprehensive discussion on the convergence of these bounds is also provided.

\begin{corollary}[A risk bound of Macro-AUC, proof in Appendix \ref{section: A}] \label{thm: sub-root AUC}
    Assume that the loss function $L$ satisfies Assumption \ref{thm:assump2}, and $\hat{\tilde{h}}$ satisfies $P _m(L _{\hat{\tilde{h}}}) = \inf _{\tilde{h} \in \mathcal{\tilde{H}}} P(L_{\tilde{h}})$. Assume: \\
    \begin{align*}
        \Phi (r) \geq B \mu \mathcal{R} \{ \tilde{h} \in \mathcal{\tilde{H}}, \mu ^2 \eE(\tilde{h}_k - \tilde{h}_k^*)^2 \leq r \},
    \end{align*}
    then for every $\tilde{h} \in \mathcal{\tilde{H}}$, with probability at least $1 - e^{-t}$,
    \begin{align*}
        \pP (L_{\hat{h}} - L_{h^*}) \leq \frac{c_1}{B} r^* +c_2\frac{t}{K},
    \end{align*}
    where $c_1 = 704$, $c_2 = (26B + 22)c $, $c = \frac{5^2}{4^2} 
\frac{1}{\tilde{n}} \sum_{k \in [K]} \frac{1}{\tau_k} $. We can notice that $c_2$ is $O(\frac{1}{\tilde{n}} \sum_k \frac{1}{\tau_k})$.
\end{corollary}

\begin{corollary}[Linear case excess risk bound of Macro-AUC, proof in Appendix \ref{section: A}] 
\label{thm: linear comput AUC} Assume that $\sup _{\vx \in \mathcal{X}} $ $ \|\vx\|_2^2 \leq \bar{M}^2,\bar{M} > 0$. The hypothesis $\mathcal{\tilde{H}}=\{ \tilde{h},\tilde{h}=(\tilde{h}_1,\dots,\tilde{h}_K), ~\tilde{h}_k = \theta_k^T \vx, ~\|\theta_k\|_2 \leq \widetilde{M} \}$. And the loss function $L$ satisfies Assumption \ref{thm:assump2}, $C$ is a constant about $B, \mu $. Then with probability at least $1 - e^{-t}$, 
\begin{align} \label{eq:linear eq AUC}
    P (L_{\hat{h}} - L_{h^*}) \leq C_{B,\mu} (r^* + \frac{C'_{\tau_1,\tau_2,...\tau_K}}{K} \frac{ t}{\tilde{n}}),
\end{align}
where
\begin{align} \label{eq:linear r AUC}
    r^* \leq 2 \sum_k \min_{d_k \geq 0} ( \frac{1}{\bar{M}^2} \cdot \frac{d_k}{\tilde{n} K} \frac{1}{\tau_k} + \widetilde{M} \sqrt{\frac{1}{\tilde{n} K} \cdot \frac{1}{\tau_k} \sum_{l > d} \widetilde{\lambda}_{l}^2 } ),
\end{align}
where $d$ is the division of singular values of matrix $\Theta$. 
\end{corollary}

\begin{proposition} \label{pro: kernel proof1}
\begin{align*}
    \sum_k \sqrt{\frac{\chi_f(G_k)}{n K} \sum_{l > d_k^*} \lambda_{kl}} \leqone \sqrt{\frac{1}{n K}} \cdot \sqrt{\sum_k \chi_f(G_k)} \cdot \sqrt{\sum_k \sum_{l > d^*_k} \lambda_{kl}}
\end{align*}
\text{\ding{172}} is due to the Cauchy Schwartz inequality, i.e., $\vx \cdot \vy \leq \|\vx\|_2 \cdot \|\vy\|_2$. Let $x_i = \chi_f(G_k)$, $y_i = \sum_{l >d_k^*} \lambda_{kl}$. 
\end{proposition}

Many combinatorial kernels are of finite rank, such as the combination of additive and multiplicative kernels, and can create finite-dimensional feature space in an abounded way. Moreover, although some kernels are infinite dimensional, within a certain error range, finite kernels can be approximated by truncation on finite data points, such as Laplacian Kernel, and Gaussian Kernel. 
\subsubsection{Some generalization analysis of linear hypothesis}
\label{sec-app:macro-auc-results}
Next, we will analyze in detail the convergence rate of the generalization boundary in linear space (Corollary \ref{thm: linear comput AUC}).

With inequality \eqref{eq:linear eq AUC}, we can notice that $P(L_{\hat{h}} - L_{h^*}) = O(r^*) + O(\frac{1}{K \tilde{n}}) $, similarly, $r^* \in [O(\frac{1}{\tilde{n}}), O(\sqrt{\frac{1}{\tilde{n}}})]$ in inequality \eqref{eq:linear r AUC}. Then we analyze $r^*$, i.e., $\frac{1}{\bar{M}^2} \cdot \frac{d'}{\tilde{n}} \sum_{k \in [K]} \frac{1}{\tau_k} + \widetilde{M} \sqrt{\frac{1}{\tilde{n}} \sum_{k \in [K]} \frac{1}{\tau_k} \sum_{l > d'} \widetilde{\lambda}_l^2 } $, by the value of $d'$, where $d' = \argmin_{d \geq 0} (\frac{1}{\bar{M}^2} \cdot \frac{d}{\tilde{n}} \sum_{k \in [K]} \frac{1}{\tau_k} + \widetilde{M} \sqrt{\frac{1}{\tilde{n}} \sum_{k \in [K]} \frac{1}{\tau_k} \sum_{l > d} \widetilde{\lambda}_l^2 }) $. 
\begin{enumerate}[(1)]
    \item $d' = 0$, in this case, then
    \begin{align*}
        r^* \leq  \widetilde{M} \sqrt{\frac{1}{\tilde{n}} \sum_{k \in [K]} \frac{1}{\tau_k} \sum_{l > 0} \widetilde{\lambda}_l^2 } 
        \leqone  \widetilde{M} M (\sqrt{\frac{1}{\tilde{n}} \sum_{l>0} \widetilde{\lambda}_l^2 }),
    \end{align*}
    where $M = \sqrt{\sum_{k \in [K]} \frac{1}{\tau_k} }$, thus we can get the inequality \text{\ding{172}}. And we can notice that $r^* = O(\sqrt{\frac{1}{\tilde{n}}})$, which is the worst case scenario for the convergence of our generalization bounds. But we know that the generalization bound of GRC analysis is also $\sqrt{\frac{1}{\tilde{n}}}$. This at least shows that our convergence rate is not worse than GRC.

    \item $r(\Theta)$ is finite, where $r(\Theta)$ refers to the rank of 
matrix $\Theta$.
    then there exists $ d' < \infty $ for $\sum_{l > 0} \widetilde{\lambda}_l^2 = 0 $, then
    \begin{align*}
        r^* \leq  \frac{1}{\bar{M}^2} \cdot \frac{d'}{\tilde{n}} \sum_{k \in [K]} \frac{1}{\tau_k}
        \leqone  \frac{M^2}{\bar{M}^2} \cdot \frac{d'}{\tilde{n}},
    \end{align*}
where $d' = \argmin_{d>0} \frac{1}{\bar{M}^2} \cdot \frac{d}{\tilde{n}} $, \text{\ding{172}} is due to $M = \sqrt{\sum_{k \in [K]} \frac{1}{\tau_k} }$ (we donate). We can notice that $r^* = O(\frac{d'}{\tilde{n}})$. If $r(\Theta) \leq $, i.e., $d' \leq a$, then $r^* = O(\frac{a}{\tilde{n}})$, where $a$ is a constant. 

    \item The eigenvalues of the SVD decomposition of the $\Theta$ matrix decay exponentially. 
    In this case, we have $\sum_{l > d'} \widetilde{\lambda}_l^2 = O(e^{-d'})$. By setting the truncation threshold at $d' = \log \tilde{n}$,
    then 
    \begin{align*}
    r^* \leq & (\sum_{k \in [K]} \frac{1}{\tau_k}) \cdot \frac{1}{\bar{M}^2} \cdot \frac{\log \tilde{n}}{\tilde{n}}  + \widetilde{M} \sqrt{\frac{1}{\tilde{n}} \sum_{k \in [K]} \frac{1}{\tau_k} \sum_{l>d'} \widetilde{\lambda}_l^2 } \\
    \leq & \frac{M^2}{\bar{M}^2} \cdot \frac{\log \tilde{n}}{\tilde{n}} + \widetilde{M} M \sqrt{\sum_{l > d'} \frac{C}{\tilde{n}^2} }  \\ 
    = & \frac{M^2}{\bar{M}^2} \cdot \frac{\log \tilde{n}}{\tilde{n}} + \widetilde{M} M \frac{\sqrt{C}}{\tilde{n}},
    \end{align*}
    where $C$ is a constant. \text{\ding{172}} is due to the known condition and some simple deflating method used earlier. Through the above analysis, we can see that $r^* = O(\frac{\log \tilde{n}}{\tilde{n}})$.

\end{enumerate}

\subsection{Other Applications}
It can be observed that the Macro-AUC optimization problem in Multi-Class Learning (MCL) \cite{21multiclass} is a special case of the aforementioned applications. Our focus remains on the same mapping function $h$, so the theoretical results mentioned above still hold in multi-class learning.
Moreover, another related application is Area Under the Uplift Curve (AUUC) - Maximization \cite{aminiKDD2021uplift_Modeling}, which has a conversion relationship with Macro-AUC Optimization.
We provide a detailed analysis of this application and present main results there (see details in Appendix \ref{pro:AUUC-all-appendix}).

\subsection{Area Under the Uplift Curve (AUUC) - Maximization}
\label{pro:AUUC-all-appendix}
\subsubsection{Problem setting}
Let $\mathcal{X} \in \sR^d$, $\vx \in \mathcal{X}$ be a feature vector, and output $Y \in \{0,1\}$. We need to introduce treatment variables $G=\{T,C\}$ that indicate whether ($g = T$) or ($g = C$) not each individual received treatment. Assume a dataset $(\vx_{ki},y_{ki},g_{ki}) \overset{i.i.d.}{\sim} D_{\mathcal{X},Y,G}; \mathcal{X} \bot G$. To simplify the representation of $S$, we use $S^g$ to represent a subset of $S$, i.e., $S^g$ can be $S^T$ or $S^C$. Then we can describe the setting: given dataset $S = \{S^C, S^T\}$, where $S^g = \{ S^g_1,S^g_2,...,S^g_K \}$. For each $k \in [K]$, $S^g_k = \{ 
(\vx_{ki}, y_{ki}, g) \}_{i=1}^{m^g_k}$. Also for every $k \in [K]$, $m_k = m^C_k + m^T_k$, $\sum_{k \in [K]} m_k = \sum_{k \in [K]} m_k^C + \sum_{k \in [K]} m_k^T = n^C +n^T = N$. The goal is to learn a mapping function $h= (h_1,h_2,...,h_K)$. 


\subsubsection{Some theory results}
\begin{definition}[AUUC-max in multi-tasks learning] \label{def:multi own2 auuc} For \( k \in [K] \), define \( h_k(S_k^T, \frac{p}{100}m_k^T) \) and \( h_k(S_k^C, \frac{p}{100}m_k^C) \) as the first \( p \) percentiles of \( S_k^T \) and \( S_k^C \), respectively, when arranged according to the predictions of each \( h_k \). Furthermore, for \( S^T \) and \( S^C \), we have \( h(S^g, \frac{p}{100} n^g) = \sum_{k \in K} h_k(S_k^g, \frac{p}{100} m_k^g) \). 
Then empirical AUUC of $h$ on $S$ can be defined as 
\begin{align*}
    \widehat{AUUC} (h,S) =  \frac{1}{K} \sum_{k=1}^K \int_{0}^{1} V(h_k,\vx) d \vx 
    \approx  \frac{1}{K} \sum_{k=1}^K \sum_{p = 1}^{100} V(h_k,\frac{p}{100}),
\end{align*}
where $V(h_k,\frac{p}{100})$ satisfies the following: 
\begin{align*}
     V(h_k,\frac{p}{100}) = \frac{1}{m_k^T} \sum_{i^T \in h_k(S_k^T, \frac{p}{100}m_k^T) } y_{k i^T} - \frac{1}{m_k^C} \sum_{i^C \in h_k(S_k^C, \frac{p}{100}m_k^C) } y_{k i^C}.
\end{align*}
Moreover, the expected AUUC of $h$ is $\eE_S [\widehat{AUUC}(h,S)]$.
\end{definition}

Since we use AUUC as the evaluation metric and bipartite ranking as the algorithm training data, we need to establish a relationship between them using the following proposition. 
\begin{proposition} \textnormal{\textbf{(The relationship between AUUC-max and the bipartite ranking loss in multitask learning.)}}  \label{pro:multi own2 ranking auuc} The definition of $\widehat{AUUC}(h,S) $ and $AUUC$ in Definition \ref{def:multi own2 auuc}. Then $AUUC$ and ranking loss satisfied the following : 
\begin{align*}
     AUUC (h)  \geq \Gamma^{T,C} -  \left( \Lambda^T \eE_{S^T} [ \hat{R}(h,S^T) ] + \Lambda^C \eE_{S^C} [ \hat{R}(h,\widetilde{S}^C) ] \right),
\end{align*}
where $\Gamma = \eE_{S} [\frac{1}{K} \sum_{k=1}^K (\bar{y}_k^T - \frac{1}{2} ((\bar{y}_k^T)^2 + (\bar{y}_k^C)^2) )]$, $\Lambda^g = \sum_{k=1}^K \bar{y}_k^g (1-\bar{y}_k^g) $, $\bar{y}_k^g = \eE [Y_k | G=g]$.     
\end{proposition}


\begin{proof} From Definition \ref{def:multi own2 auuc}, we know the empirical AUUC of $h$, then 
\begin{align*}
    \widehat{AUUC} (h,S) = & \frac{1}{K} \sum_{k=1}^K \int_{0}^{1} V(h_k,\vx) d \vx 
    \eqone   \frac{1}{K} \sum_{k=1}^K \int_{0}^{1} ( F_{h_k}^{S^T_k}(\vx) - F_{h_k}^{S^C_k}(\vx) ) d \vx \\
    = & \frac{1}{K} \sum_{k=1}^K \left( \int_{0}^{1} F_{h_k}^{S^T_k}(\vx) d \vx + \int_{0}^{1} F_{h_k}^{S^C_k}(\vx) d \vx \right) \\
    \eqtwo & \frac{1}{K} \sum_{k \in [K]} ( \bar{y}_k^T (1-\bar{y}_k^T) \cdot AUC(h_k, S_k^T) + \frac{(\bar{y}_k^T)^2}{2}  
    \\ & - \bar{y}_k^C (1-\bar{y}_k^C) \cdot AUC(h_k, S_k^C) -  \frac{(\bar{y}_k^C)^2}{2} ) \\
    \eqthree & \frac{1}{K} \sum_{k \in [K]} ( \bar{y}_k^T (1-\bar{y}_k^T) \cdot AUC(h_k, S_k^T) + \frac{(\bar{y}_k^T)^2}{2}  
    \\ & - \bar{y}_k^C (1-\bar{y}_k^C) \cdot (1 - AUC(h_k, \widetilde{S}_k^C))  - \frac{(\bar{y}_k^C)^2}{2} ) \\
    \eqfour & \frac{1}{K} \sum_{k \in [K]} ( \bar{y}_k^T (1-\bar{y}_k^T) \cdot (1 - \hat{R}_k (h_k, S_k^T)) + \frac{(\bar{y}_k^T)^2}{2}  
    \\ & -   \bar{y}_k^C (1-\bar{y}_k^C) \cdot (1 - (1 - \hat{R}_k (h_k, \widetilde{S}_k^C)) - \frac{(\bar{y}_k^C)^2}{2} ) \\
    = & \frac{1}{K} \sum_{k = 1}^K  ( \underbrace{ \bar{y}_k^T - \frac{1}{2} ((\bar{y}_k^T)^2 + (\bar{y}_k^C)^2) }_{\hat{\gamma}_k^{T,C}}  )  
    -   \frac{1}{K} \sum_{k=1}^K ( \underbrace{ \bar{y}_k^T (1-\bar{y}_k^T) }_{\lambda_k^T} \hat{R}_k(h_k,S_k^T)  
    \\ & +   \underbrace{ \bar{y}_k^C (1-\bar{y}_k^C) }_{\lambda_k^C} \hat{R}_k(h_k,\widetilde{S}_k^C) ) \\
    \geqfive & \hat{\Gamma}^{T,C} -  \left( \Lambda^T \hat{R}(h,S^T) + \Lambda^C \hat{R}(h,\widetilde{S}^C) \right).
\end{align*}
\text{\ding{172}} is due to \cite{aminiKDD2021uplift_Modeling,KDD2011_Ffunction}, i.e.,  $F_{h_k}^{S^T_k}(\vx), F_{h_k}^{S^C_k}(\vx)$ can be induced by $h$, and $V(h_k,\vx) = F_{h_k}^{S^T_k}(x) - F_{h_k}^{S^C_k}(\vx)$. \text{\ding{173}} is due to \cite{aminiKDD2021uplift_Modeling,KDD2011_ginifunction}, i.e., a relationship between $F_{h_k}^D$, $Gini(h, D)$ and $AUC$, where D represents a dataset. Then we can get $\int_{0}^{1} F_{h_k}^D(\vx) d \vx = \bar{y}_D (1-\bar{y}_D) \cdot AUC (h,D) + \frac{(\bar{y}_D)^2}{2} $, and then substitute it to \text{\ding{172}}. In \text{\ding{174}}, we revert labels in $S^C_k$ for $\forall k \in [K]$, i.e., $\widetilde{S}_k^C = \{\vx_{ki},1-y_{ki},C \}_{i=1}^{m_k^C}$. Thus $AUC(h_k,S_k^C) = 1 - AUC (h_k, \widetilde{S_k^C})$ and substitute it to \text{\ding{173}}. Moreover, \text{\ding{175}} is due to the relationship between $AUC$ and empirical ranking risk, i.e., $AUC(h_k,D) = 1-\hat{R}_k(h_k,D)$, where $\hat{R}_k (h_k,D) = \frac{1}{|D^+||D^-|} \sum_{(\vx_{ki},\vx_{kj}) \in D^+ \times D^-} [\! [h_k(\vx_{ki}) \leq h_k(\vx_{kj})]\!] $. The last inequality \text{\ding{176}} is due to $\sum_{i \in [n]} a_i b_i \leq \sum{i \in [n]} a_i \cdot \sum{i \in [n]} b_i$. 
Finally, we take expectations from both sides, then
\begin{align*}
    AUUC (h) \geq \Gamma^{T,C} -  \left( \Lambda^T \eE_{S^T} [ \hat{R}(h,S^T) ] + \Lambda^C \eE_{S^C} [ \hat{R}(h,\widetilde{S}^C) ] \right).
\end{align*}
\end{proof}


Then we can define our empirical risk of $h$ as follows:
\begin{align*}
    \hat{R}_S = & \hat{R}_{S^T} + \hat{R}_{S^C}, \\
    \hat{R}_{S^g} = & \frac{1}{K} \sum_{k =1}^K \frac{1}{m_k^g} \sum_{i=1}^{m_k^g} L(\vx_{ki},y_{ki},g,h_k), g \in \{ T,C \}. 
\end{align*}
The expected risk of $h$ satisfies $R(h)=\eE _S[\hat{h}]$. And according to the definition of LFRC in Definition \ref{def: LFRC}, we can define the empirical LFRC of $S^g$ for $g \in \{ T,C \}$ as
\begin{align*}
    \hat{\mathcal{R}}_{S^g}(\mathcal{H},r) =  \frac{1}{K} \eE_{\zeta} \left[ \sup_{h \in \mathcal{H},\mathrm{var}(h) \leq r} \sum_{k \in [K]} \frac{1}{m_k^g} \sum_{j \in J_k^g} \omega_{kj} \sum_{i \in I_{kj}^g} \zeta_{ki} h_k(\vx_{ki}) \right].
\end{align*}

\begin{theorem}[The base theorem of AUUC-max in multi-tasks learning] \label{thm:multi own2 2.1 base} For each $t > 0$, with probability at least $1-2e^{-t}$,
\begin{align*}
    AUUC (h) \geq &  \Gamma^{T,C} - ( \Lambda^T \hat{R}(L \circ h, S^T) + \Lambda^C \hat{R}(L \circ h, S^C) )  \\ & -  R_{S^T,S^C} (H,S^T,S^C) - [(\frac{2}{3} + \frac{1}{\alpha^T} ) c^T + (\frac{2}{3} + \frac{1}{\alpha^C} )c^C] \frac{t}{K} ,
\end{align*}

where $R_{S^T,S^C} (H,S^T,S^C)$ is related to the LFRC of $S^T$ and $S^C$, i.e., $R_{S^T,S^C} (H,S^T,S^C)$ = $2(1+\alpha^T) \mathcal{R}_{S^T}(\mathcal{H},r) + \sqrt{\frac{2c^T rt}{K}} + 2(1+\alpha^C) \mathcal{R}_{S^C}(\mathcal{H},r) + \sqrt{\frac{2c^C rt}{K}}$. Additionally, $c^g = \frac{5^2}{4^2} 
\sum_{k \in [K]} \frac{\chi_{h,g}(G)}{m_k^g} $ for $g \in \{ T,C \}$.    
\end{theorem}

\section{Experimental Results Supplement} \label{sec-app:experiment appendix}

\begin{table}
\caption{The estimated values (mean $\pm$ std) of 
$r^*$ in these datasets.}
\label{table: estimate_var}
\centering
\begin{threeparttable} 
\begin{tabular}{lcc}
    \toprule
    Dataset  & $r^*$   \\
    \midrule
    Emotions   & 0.234 $\times$ 1e-3 $\pm$ 0.322 $\times$ 1e-4  \\
    CAL500    & 0.871 $\times$ 1e-2 $\pm$ 0.127 $\times$ 1e-3 \\
    Image    & 0.290 $\times$ 1e-5 $\pm$ 0.624 $\times$ 1e-6 \\
    Scene    &  0.143 $\times$ 1e-4 $\pm$ 0.499 $\times$ 1e-6  \\
    Yeast    & 0.160 $\times$ 1e-2 $\pm$ 0.102 $\times$ 1e-3 \\
    Corel5k       &  0.762 $\times$ 1e-1 $\pm$ 0.183 $\times$ 1e-2 \\
    Rcv1subset1     & 0.297 $\times$ 1e-2 $\pm$ 0.524 $\times$ 1e-3 \\
    Bibtex   &  0.601 $\times$ 1e-3 $\pm$ 0.369 $\times$ 1e-4 \\
    Delicious   & 0.131 $\times$ 1e-2 $\pm$ 0.441 $\times$ 1e-4 \\
\bottomrule
\end{tabular}
\end{threeparttable}
\end{table}

Here, some experimental details from the main text are supplemented (see Table~\ref{table:variables in bounds}), and the tables of experimental results mentioned in the main text are provided (see Table~\ref{table: upper bound of pa}, Table~\ref{table: estimate_var}).

\subsection{Theoretical supplement regarding the experiment}

Because our results differ from the form presented in \cite{wu2023macro-auc}, the following derivation is needed for a more fair comparison. 
$h^*$ represents the minimum value of the expected risk, using techniques similar to \cite{mohri2018foundations}, then for arbitrarily $ \epsilon >0$, there exists $ h_{\epsilon}$, such that $P(L_{h_\epsilon}) \leq P(L_{h^*}) + \epsilon$. Then
\begin{align*}
    P(L_{\hat{h}} - L_{h^*}) & = P(L_{\hat{h}}) - P(L_{h_\epsilon}) + P(L_{h_\epsilon}) - P(L_{h^*}) 
     \leq P(L_{\hat{h}}) - P(L_{h_\epsilon}) + \epsilon \\
    & = P(L_{\hat{h}}) - P_m(L_{\hat{h}}) + P_m(L_{\hat{h}}) - P(L_{h_\epsilon}) +\epsilon \\
    & \leqone P(L_{\hat{h}}) - P_m(L_{\hat{h}}) + P_m(L_{h_\epsilon}) - P(L_{h_\epsilon}) +\epsilon 
     \leq 2 \sup_{h \in \mathcal{H}} |P(L_h) - P_m (L_h)| +\epsilon ,
\end{align*}
where $P(L_{\hat{h}} - L_{h^*})$ can be seen in Corollary \ref{thm : Lipschitz bound loss space}. \text{\ding{172}} is due to $P_m(L_{\hat{h}}) \leq P_m(L_{h_\epsilon})$. The two bounds we want to compare are as follows:
\begin{align} \label{eq: experiment our}
    P(L_{\hat{h}} - L_{h^*}) \leq 704 \mu r^* +  \frac{75}{K} \sum_{k \in [K]} \frac{1}{\tau_k} \cdot \frac{t}{\tilde{n}},
\end{align}
where 
\begin{align} \label{eqappendix:the value of r}
    r^* \leq 2 \min_{d \geq 0, d \in \sN} ( \frac{1}{\bar{M}^2} \cdot \frac{d}{\tilde{n} K} \sum_{k \in [K]} \frac{1}{\tau_k} + \widetilde{M} \sum_{k \in [K]}\sqrt{\frac{1}{\tilde{n} K}  \frac{1}{\tau_k} \sum_{l > d} \widetilde{\lambda}_l^2 } ),
\end{align}
\begin{align} \label{eq:experiment old}
    P(L_{\hat{h}} - L_{h^*}) \leq 2 \left[ \frac{4 \mu \bar{M} \widetilde{M}}{\sqrt{\tilde{n}}} \left(\frac{1}{K} \sum_{k \in [K]}\sqrt{\frac{1}{\tau_k}}\right) + 3 \sqrt{\frac{\log 2 +t}{2 \tilde{n}}} \left(\sqrt{\frac{1}{K} \sum_{k \in [K]}\frac{1}{\tau_k}}\right) \right], 
\end{align}
thus we should calculate and compare the right-hand sides of Eq. \eqref{eq: experiment our} and Eq. \eqref{eq:experiment old}. If the result of the former is smaller than that of the latter, we can approximately conclude that our bound is tighter.

\subsection{Experimental details}
\label{section:expdetails}
Here, we will give a detailed explanation of the datasets used in the experiment and some computational issues such as parameters (see details in Table~\ref{table:variables in bounds}). 

\textbf{Datasets}. The datasets are about multi-label data. Information such as the sample size and label number of the dataset is shown in Table \ref{table: upper bound of pa}. Moreover, the datasets are available at http://mulan.sourceforge.net/datasets-mlc.html and http://palm.seu.edu.cn/zhangml/.

\textbf{Experiments setting and computing bounds}. We choose stochastic gradient descent (SGD) as the optimizer,and the hyperparameter of the weight decay is selected in $\{ 10^{-4},10^{-3},10^{-2},10^{-1} \}$. We use linear models ($f=W^Tx, W=(w_1,\dots,w_K)^T$) on the tabular benchmark datasets, which adopt a learning rate of $0.05$ and $300$ epochs. We calculate the bound of the pa algorithm (see \cite{wu2023macro-auc} for details). 
Eq. \eqref{eq: experiment our}, \eqref{eq:experiment old} are risk bounds of the pa algorithm. (see parameter details in Table~\ref{table:variables in bounds}).

The experiments are conducted on a server with $6$ GPUs (GeForce RTX 3090) with $256$G RAM and $2$T storage and be run with $1$ GPU.

\begin{table}
\caption{Explanation of the variables in the risk bounds. 
}
\label{table:variables in bounds}
\centering
\begin{threeparttable} 
\begin{tabular}{p{3cm}p{4cm}p{5cm}}
\toprule
Variable symbols  & The meaning of variables & Value or the conditions to be satisfied    \\
\midrule
$\mu$ & Lipschitz constant for loss function & $1$ (hinge loss) \\
$r^*$ & The fixed point of $\Phi(r)$\tnote{1} & satisfying Eq. \eqref{eqappendix:the value of r} \\
$K$ & number of labels & value details in Table \ref{table: upper bound of pa} (e.g., for Emotions, $K=6$) \\
$\tau_k$ & the imbalance level factor \tnote{2}& $\frac{\min \{ \tilde{n}_k^+, \tilde{n}_k^- \}}{n}$ \\
$\tilde{n}$ & training data size  & using 3-fold cross-validation, training data : testing data $= 2:1$ \\
$t$ & related to the confidence level of risk bounds  & $\ln 100$ (owing to setting $1-e^{-t}=0.99$) \\
$\bar{M}$ & the norm constraint of the input space & $\bar{M} \geq \sup_x \| x \|_2$ \\
$\widetilde{M}$ & the norm constraint of the weight vector & $\widetilde{M} \geq \sup_k \|w_k\|_2$ \\
$d$ & the parameter for partitioning and truncating singular values\tnote{3} & $ \min\{ D, K\} \times \text{rate}$ \\
$\widetilde{\lambda}^2_{l}$ & the $l$-th largest singular value \tnote{4} of the matrix $W$  & $W=\sum_{l = 1}u_lv_l^T \widetilde{\lambda}_{l}$ \\
\bottomrule
\end{tabular}

\begin{tablenotes}
    \footnotesize
    \item[1] $\Phi(r)$ is a sub-root function, which can be seen in Corollary \ref{thm : Lipschitz bound loss space}.
    \item[2] The details of $\tau_k$ can be found in problem setup in Section \ref{section: applications}.
    \item[3] Use a method similar to that in \cite{experiment23li} to estimate the value of $d$. The $\text{rate} \in [0,1]$, with values (1,1,1,1,1,1,1,1,1) for datasets (emotions $\sim$ delicious), respectively. $D$ is the input feature. 
    \item[4] ${\{\widetilde{\lambda}_l\}}_{l = 1}^{\infty}$ are the singular values of $W$ and are sorted in a nonincreasing order.
\end{tablenotes}

\end{threeparttable}
\end{table}

\section{Supplementary Discussion} \label{section: discussion}

\begin{table}[t]
\caption{Comparison of the main results about $n$ and $K$ \tnote{1}} 
\label{table: full_discuss_nk}
\begin{center}
\begin{small}
\begin{threeparttable} 
\begin{tabular}{lll}
\toprule
Reference  & Setting & The convergence rate of the results 
\\
\midrule
 \cite{Bartlett_2005} &  single-task, iid & $O(\frac{\log n}{n})$  \\
 \cite{ralaivola2015entropy} & single-task,  single-graph dependent  & --   \\
 \cite{yousefi18} & multi-task, iid & $O(\frac{\log n}{n K})$  \\
 \cite{watkins2023optimistic} & multi-task representation, iid & $O(\frac{\log n}{n K})$  \\
 \cite{wu2023macro-auc} & multi-task, multi-graph dependent & $O(\frac{1}{\sqrt{n}} \max\{\frac{1}{K} \sum_k \sqrt{\chi_f(G_k)}, \sqrt{\frac{1}{K} \sum_k \chi_f(G_k)} \})$  \\
 ours & multi-task, multi-graph dependent & $O(\sum_k \frac{\mathrm{Rank}(\kappa_k) \chi_f(G_k)}{n K})$ \\
\bottomrule
\end{tabular}
\begin{tablenotes}
    \footnotesize
    \item[1] $n$ is the number of samples for each task, $K$ is the number of tasks. 
   
\end{tablenotes}
\end{threeparttable}
\end{small}
\end{center}
\end{table}

\textbf{Comparison of technique and theoretical results with those in the previous literature} (see Table~\ref{table: technique_results_discuss}, \ref{table: full_discuss_nk}). 
Compared with \cite{ralaivola2015entropy}, technically, we prove its hypothesis of fast-rate convergence, as stated: "Note this result is only the starting point of a wealth of results that may be obtained using the concentration inequalities studied here. In particular, it might be possible to study how arguments based on star hulls and sub-root functions may help us to get fast-rate-like results..." 
Compared with i.i.d. case \cite{Bartlett_2005,yousefi18,watkins2023optimistic}, we propose new techniques by graph theory to handle multi-graph dependent data and new Bennett-type and Talagrand-type inequality, and Local fractional Rademacher complexity (LFRC), different from the ones in i.i.d.(e.g., local Rademacher complexity (LRC)). 

By comparing the convergence rates of our theoretical bounds with those in other literature under the multi-task setting, it can be found that regarding the dependence on the number of samples $n$ for each task, we have obtained a faster convergence rate for multi-graph dependent variables than that in \cite{wu2023macro-auc}, and it is on par with the fast convergence rate of the i.i.d. bounds. 
Regarding the dependence on the number of tasks $K$, our result shows a dependence of $O(\sum_k \frac{\mathrm{Rank}(\kappa_k) \chi_f(G_k)}{n K})$, while under the i.i.d. assumption, the best-known dependence on $K$ is $O(\frac{1}{K})$ \cite{yousefi18,watkins2023optimistic}. Since the processing of multi-graph dependent variables introduces additional $K$-related terms, our dependence on $K$ is not tight, which is our limitation, left as future work. 



\textbf{The tightness and generality of Bennett's inequality}. The Bennett inequality (Theorem~\ref{thm:bennett_inequality},~\ref{thm:bennett_inequality_refined} (Appendix~\ref{sec-app:a_special_bennett_inequality}) ) we propose is tight and general. 
Especially, in Theorem \ref{thm:bennett_inequality}, 
our Bennett inequality is for multi-graph dependent variables. 
In particular, the result of \cite{ralaivola2015entropy} for the (single) graph-dependent case is a special instance of our inequality (i.e., \( K=1 \)). In addition, in Theorem \ref{thm: bennett inequality lower bound} (Appendix~\ref{sec-app:supplement-bennett-lower}), we complement Theorem \ref{thm:bennett_inequality} by providing the two-sided constraint of the Bennett inequality, i.e., $\pP(|Z-\eE Z| \geq t) \leq 2 \exp (-\frac{v}{\sum_{k \in [K]} \chi_f(G_k)} \varphi(\frac{4t}{5v}))$. This helps explain the tightness of our bound to some extent, as the coefficients are the same. Besides, \cite{Bartlett_2005} (i.i.d.) is also a special case of ours; however, our result does not encompass its conclusion, though there is a constant (i.e., \( \frac{4}{5} \)). 

To address these scenarios, we provide a specific Bennett inequality (i.e., \( \omega=1 \), detailed in Appendix \ref{sec-app:a_special_bennett_inequality}), which includes \cite{Bartlett_2005}. However, due to the coarser scaling in its derivation, this inequality may not be optimal, particularly when \( K \neq 1 \), resulting in relatively loose outcomes. Our rigorous calculations indicate that this result is only somewhat close to Theorem \ref{thm:bennett_inequality}, which remains tighter.

Thus, we need to employ scaling or function approximation techniques to optimize the Bennett inequality for non-i.i.d. variables, enhancing its generality and tightness, especially in case \( w=1 \) to ensure continuity of the inequality.

\textbf{The tightness of our risk bound.} 
Our risk bound has a tight dependence on the sample size $n$, while its dependence on the number of tasks $K$ is weaker than the optimal solution under the i.i.d. assumption. However, for multi-graph dependent variables, the dependence on $K$ is the same as that in \cite{wu2023macro-auc} (see details in Appendix~\ref{section: discussion}). 
Furthermore, we comput the upper and lower bounds (see in Proposition \ref{thm:kernel upper bound} (see Section~\ref{sec-mtl:risk-bound-for-mtl}), \ref{thm:linear upper bound} (see Appendix~\ref{sec-app:supplemental-risk-bounds})) for LFRC, which indicates that our bounds are relatively tight and the experimental results corroborate our theoretical findings (see 
 Section~\ref{section: C} for details). However, for the lower bound, we only provided a somewhat vague estimate and did not conduct a detailed investigation or discussion about the constant \( c \), suggesting that there is still room for improvement in our bound values. Moreover, in deriving the generalization bounds, we repeatedly utilize the Cauchy-Schwarz inequality and other coarse scalings, which results in bounds that are not sufficiently tight. These limitations highlight areas for further research in future work.

\textbf{The optimization of the fixed point \( r^* \).}
Our bounds are related to the complexity of $r^*$; however, we followed the approach of other literature by setting it as a hyperparameter in our experiments, resulting in approximate results. This has created a gap between the experimental and theoretical results. Additionally, the computed risk bound from the experiments is not non-empty (i.e., greater than 1), which is also a limitation of our study. However, compared to \cite{wu2023macro-auc}, our bounds are tighter in most datasets, and these theoretical results still provide valuable guidance for algorithm design.

\textbf{Other hypothesis space.} For Neural Networks (NN), similar theoretical results may be achieved, but non-trivial. 
Especially, in i.i.d. case, some papers have utilized classical methods such as CN \cite{cn2024nn} and RC \cite{rcnn2024} to analyze the generalization bound of NN and obtain the best convergence result of $O(\frac{1}{\sqrt{n}})$. Some studies \cite{empiricallrc19} have found that adding the LRC regularization term to NN can improve the model performance, and some other studies \cite{lrcbound21} have calculated the empirical upper bound of LRC. However, they did not further explore the techniques of LRC (such as fixed point, star-hull) under the i.i.d. assumption, and thus did not obtain a faster convergence rate with respect to $n$.
However, based on our proposed new method, this generalization bound is expected to be theoretically improved. Nevertheless, the study still faces numerous challenges \cite{18nnoverparameter}, including some optimization problems that are non-convex and have significant overparameterization phenomena. Additionally, to facilitate analysis, research \cite{CNnn19} typically assumes explicit regularization using p-norm, whereas, in practical neural network applications, regularization is often implicit, which may lead to a deviation between the assumed constraints \cite{rc18nn} and the actual application, thereby making the generalization analysis bound appear even broader. 
Despite the presence of these difficulties, We believe that our proposed new concentration inequality will have a positive impact on the generalization analysis of neural networks, and can be extended to other theoretical analysis fields, providing more possibilities for future research work.

\textbf{Other applications}. Some applications involve processing a large graph, such as when nodes in a social network need to be labeled with different labels. In this case, it is necessary to address classification problems that depend on multiple nodes \cite{11nodeclassification}, which will serve as an application of the theoretical results presented in this paper.

\textbf{Border Impacts}. This paper presents a theoretical study on Multi-Task Learning (MTL), aimed at providing theoretical support and guidance for algorithms, to facilitate their development, while without any significant societal impacts. 

\section{Additional Related Work}  \label{section: D}
\textbf{Concentration inequalities.} \cite{zhang2022generalization} delineates the evolution of concentration inequalities, continuously refining aspects such as universality, adaptability, and tightness. First, \cite{Mcolin89} introduced the McDiarmid inequality, applicable to independent random variables, establishing a theoretical framework for analyzing generalization error; however, its limitation lies in its inability to address practical problems involving dependent variables. Subsequently, \cite{jason04} proposed a concentration inequality tailored for partially dependent random variables, drawing from Hoeffding's inequality, which primarily handles graph-dependent random variables, but is constrained to functions involving summation operations. To mitigate this limitation, \cite{usunier2005generalization} further extended the study of concentration inequalities by introducing a universal concentration inequality applicable to functions of random variables of any form, thereby providing broader theoretical support for subsequent research.

Building on this foundation, \cite{ralaivola2015entropy} advanced the discourse by extending the Bennett inequality for application in LRC analysis, facilitating the exploration of generalization bounds, and maturing the application of concentration inequalities in datasets with complex dependency structures. Moreover, \cite{lampert18} considered weak dependence relationships and introduced a method to measure dependencies within unordered sets, presenting a concentration inequality applicable to any set of random variables, a contribution that significantly expands the applicability of concentration inequalities. Furthermore, \cite{zhang19mcdiarmid} proposed a concentration inequality for Lipschitz continuous functions specifically addressing tree-structured dependency graphs and demonstrated that this inequality exceeds previous studies in terms of applicability and boundary tightness, illustrating the robust adaptability of concentration inequalities within complex data environments.

Furthermore, \cite{ruiray22} provided a comprehensive theoretical enhancement to \cite{zhang19mcdiarmid} and discovered that the bounds for independent random variables were tighter than those established in \cite{jason04}. In addition, \cite{combes2024extension,tanoue2024concentration} proposes new concentrated inequalities for parameter extension and data-dependent structures of concentrated inequalities. Inspired by the concentration inequalities related to graph dependencies, the work \cite{wu2023macro-auc} proposed a concentration inequality aimed at addressing multi-label learning, subsequently deriving Rademacher complexity for analyzing generalization bounds. In light of these insights, we are inspired to build on the contributions of \cite{usunier2005generalization,ralaivola2015entropy,wu2023macro-auc} by proposing a Bennett inequality based on graph dependencies, from which we aim to derive LRC, which is expected to yield improved risk bounds compared to work  \cite{wu2023macro-auc}.

\textbf{Multi-task generalization.} 
In recent years, Multi-Task Learning (MTL) \cite{zhang2021survey} has garnered significant attention in both theoretical and applied domains, particularly in addressing challenges related to insufficient labeled data and inter-task information sharing. The theoretical analysis of MTL predominantly focuses on frameworks such as covering Numbers (CN), Rademacher Complexity (RC), and VC-dimension (VCd). For instance, \cite{ando05unableddata} proposed a framework that leverages the structural information inherent in MTL, achieving a convergence bound of $O(\frac{K}{\sqrt{n}})$ through a joint empirical risk minimization approach, where this bound is related to the number of tasks $K$. In contrast, the paper \cite{crammer12sharedhypothesis} utilized VCd to perform an in-depth analysis of shared hypotheses, generating a convergence bound of $O(\sqrt{\frac{K \log n}{n}})$, despite its somewhat limited exploration of data feature impacts.

In addition, \cite{baxter95} modeled internal representation learning as a multi-task learning scenario and acquired a convergence bound of $O(\frac{1}{\sqrt{n K}})$ using CN. Subsequently, \cite{baxter00} integrated a bias learning model with MTL, employing VCd to derive a convergence bound of $O(\sqrt{\frac{C}{n}})$. Furthermore, Maurer's previous work \cite{maurer06linear,maurer06rade}
applied RC to analyze MTL within linear hypothesis spaces, resulting in respective convergence bounds of $O(\frac{B}{\sqrt{n}})$, indicating data-dependent convergence rates.

In exploring task-relatedness, \cite{ben08} introduced the concept of inter-task connections and analyzed generalization boundaries using VC dimension techniques, achieving $O(\frac{1}{\sqrt{n}})$ results. Notably, many of the aforementioned bounds exhibit a linear dependence on the number of tasks K, which may lead to relaxation or even failure when K significantly exceeds n. To address this, \cite{juba06,kakade12} employed compression and regularization techniques, significantly tightening the convergence bounds to $O(\frac{C}{\sqrt{n}})$ and $O(\sqrt{\frac{\log \min(K,d)}{n}})$, respectively, where C is associated with the compression coefficient. Following the trajectory of regularization techniques, \cite{pontil2013excess} used trace norm regularization to attain a convergence bound of $O(\min(\sqrt{\frac{C}{n}}, \sqrt{\frac{\log nK}{nK}}) + \sqrt{\frac{1}{nK}})$.

Furthermore, studies \cite{pentina2015multi,zhang15multi,maurer2016benefit} have utilized CN, concentration inequalities, Gaussian averages, and other analyses to examine various algorithmic applications of MTL, achieving convergence rates of the order of $O(\frac{1}{\sqrt{n K}})$. To further enhance the convergence rates, \cite{yousefi18,watkins2023optimistic} employed LRC \cite{Bartlett_2005} to analyze theoretical bounds, resulting in convergence rates of $O(\frac{1}{(n K)^{\alpha}})$ ($0.5 < \alpha < 1$). Furthermore, \cite{wu2023macro-auc} proposed more comprehensive concentration inequalities, which, when combined with RC, analyzed the generalization bounds under non-independent and identically distributed (non-i.i.d.) conditions, producing a convergence behavior of $O(\frac{B}{\sqrt{n}} \max\{ \frac{1}{K} \sum_k \sqrt{\chi_f(G_k)}, \frac{1}{\sqrt{K}} \sqrt{\chi_f(G_k)} \})$ and highlighting the potential implications of this research in practical applications. Finally, to improve MTL efficacy, \cite{qi2024multimatch} introduced the MultiMatch method, which significantly improved experimental outcomes by integrating MTL with high-quality pseudo-label generation, demonstrating a convergence boundary of $O(\frac{d}{\sqrt{n}})$. Therefore, following works \cite{yousefi18,watkins2023optimistic,wu2023macro-auc}, we continue to explore this trajectory employing LRC in conjunction with Hoeffding's inequality, aiming to derive tighter convergence bounds than those presented in \cite{wu2023macro-auc}.

\textbf{Approaches to dependent data.} The research on methods for handling dependent random variables focuses mainly on three strategies: mixing models, decoupling, and graphical dependence. \cite{rosenblatt56} introduce the simple boundary mixing conditions, which led to the establishment of the central limit theorem for dependent random variables, revealing that results under independently and identically distributed (i.i.d.) conditions can encompass Hoeffding inequality. Subsequent work \cite{volkonskii59,lbragimov62,kontrovich07,strinwart09} further proposed extreme limit theorems that satisfy stronger mixing conditions, providing a wealth of theoretical results for the study of dependent random variables. The mixing coefficients within this approach allow for the quantitative characterization of the dependencies between data. However, this quantitative treatment is relatively complex and requires strong assumptions, which impose limitations on its practical applications. Furthermore, \cite{de1999decoupling} proposed a decoupling approach, which transformed dependent random variables into combinations of independent random variables. This work explored extreme limit theorems related to U-processes and summative dependent variables, subsequently leading to the introduction of more general decoupling inequalities, thus paving new avenues for subsequent research. Despite the theoretical richness of the decoupling methods, there are still practical challenges associated with their application.

In contrast to mixing conditions, the graphical dependence offers a more straightforward and intuitive method to present the dependencies among data. Research conducted in \cite{jason04} utilizing the method of graph coloring provided a qualitative perspective to understand the dependencies between random variables, significantly simplifying the description of these relationships and facilitating theoretical analysis and empirical validation. Furthermore, \cite{janson88} analyzed the convergence of summative dependent variables using graphical constructions, further corroborating the efficacy of the graphical dependence approach. In recent years, related studies \cite{jason04,usunier2005generalization,ralaivola2015entropy,wu2023macro-auc} have extensively employed graphical dependence methods to address dependent random variables, introducing various types of concentration inequalities and providing theoretical support for related algorithms. Consequently, we will also adopt the graph coloring method to analyze the generalization error of MTL in the context of dependent random variables.

\end{document}